\documentclass{article}

\usepackage{microtype}
\usepackage{graphicx}
\usepackage[font=small]{caption}
\usepackage{subcaption}
\usepackage{rotating}
\usepackage{booktabs} 
\usepackage{hyperref}
\usepackage{enumitem}

\usepackage[accepted]{icml2023}

\usepackage[utf8]{inputenc}
\usepackage[swedish,english]{babel}
\usepackage{enumerate} 

\usepackage{amsmath, amsthm, amssymb, amsfonts, mathrsfs} 
\usepackage{nicefrac}
\usepackage{mathtools}
\usepackage{bm, bbm}
\usepackage{thm-restate}

\usepackage[nameinlink, sort, compress]{cleveref}
\usepackage{letltxmacro,xparse}

\usepackage{tcolorbox}

\usepackage{lscape}
\usepackage{epstopdf}
\usepackage{graphicx}
\usepackage{bm}	%
\usepackage{multicol}
\usepackage{lipsum}
\usepackage{comment}
\usepackage{xspace}

\usepackage{tikz}
\usetikzlibrary{tikzmark,calc,decorations.pathreplacing}
\usepackage{amsmath, amsthm, amssymb, amsfonts, mathrsfs} 
\usepackage{nicefrac}
\usepackage{mathtools}
\usepackage{bm, bbm}
\usepackage{thm-restate}
\usepackage{booktabs}

\newcommand{\bB}{\ensuremath{\mathbb{B}}}

\newcommand{\bE}{\ensuremath{\mathbb{E}}}

\newcommand{\bN}{\ensuremath{\mathbb{N}}}

\newcommand{\bR}{\ensuremath{\mathbb{R}}}
\newcommand{\bS}{\ensuremath{\mathbb{S}}}

\newcommand{\bV}{\ensuremath{\mathbb{V}}}

\newcommand{\cA}{\ensuremath{\mathcal{A}}}

\newcommand{\cD}{\ensuremath{\mathcal{D}}}

\newcommand{\cL}{\ensuremath{\mathcal{L}}}

\newcommand{\cN}{\ensuremath{\mathcal{S}_{\textnormal{neg}}}}
\newcommand{\cO}{\ensuremath{\mathcal{O}}}
\newcommand{\cP}{\ensuremath{\mathcal{P}}}

\newcommand{\cT}{\ensuremath{\mathcal{T}}}

\newcommand{\cW}{\ensuremath{\mathcal{W}}}

\newcommand{\cZ}{\ensuremath{\mathcal{Z}}}

\newcommand{\sfp}{\ensuremath{p}}
\newcommand{\sfq}{\ensuremath{q}}

\newcommand{\sfs}{\ensuremath{\mathsf{s}}}

\definecolor{kth-blue}{RGB}{25,84,166}
\definecolor{kth-pink}{RGB}{216,84,151}
\definecolor{kth-gray}{RGB}{101,101,108}
\definecolor{green(pigment)}{rgb}{0.0, 0.65, 0.31}
\definecolor{blue(pigment)}{rgb}{0.2, 0.2, 0.6}

\crefdefaultlabelformat{#2\textbf{#1}#3}
\Crefname{definition}{\textbf{Definition}}{\textbf{Definitions}}
\Crefname{theorem}{\textbf{Theorem}}{\textbf{Theorems}}

\newcommand{\kldiv}{\ensuremath{D_{\textnormal{KL}}}}
\newcommand{\mi}{\ensuremath{I}}
\newcommand{\ent}{\ensuremath{H}}

\newcommand{\csim}{\ensuremath{\textnormal{sim}}}

\newcommand{\ind}[2]{#1^{(#2)}}

\newcommand{\infonce}{\texttt{InfoNCE}\xspace}
\newcommand{\ba}{\texttt{ER}\xspace}
\newcommand{\er}{\texttt{ER}\xspace}
\newcommand{\nce}{\texttt{NCE}\xspace}

\newcommand{\dimm}{\texttt{DIM}\xspace}

\newcommand{\moco}{\texttt{MoCo}\xspace}
\newcommand{\simclr}{\texttt{SimCLR}\xspace}

\newcommand{\swav}{\texttt{SwAV}\xspace}
\newcommand{\byol}{\texttt{BYOL}\xspace}
\newcommand{\cmc}{\texttt{CMC}\xspace}
\newcommand{\deepcluster}{\texttt{DeepCluster}\xspace}
\newcommand{\dino}{\texttt{DINO}\xspace}
\newcommand{\ir}{\texttt{IR}\xspace}

\newcommand{\lcmc}{\ensuremath{\cL_\textnormal{CMC}}}

\newcommand{\lbyol}{\ensuremath{\cL_\textnormal{BYOL}}}

\newcommand{\lswav}{\ensuremath{\cL_\textnormal{SwAV}}}
\newcommand{\lswavone}{\ensuremath{\cL_\textnormal{SwAV,1}}}
\newcommand{\lswavtwo}{\ensuremath{\cL_\textnormal{SwAV,2}}}
\newcommand{\ldeepcluster}{\ensuremath{\cL_\textnormal{DeepCluster}}}
\newcommand{\ldino}{\ensuremath{\cL_\textnormal{DINO}}}

\DeclareMathOperator*{\argmax}{arg\,max}

\DeclareMathOperator*{\arginf}{arg\,inf}

\theoremstyle{plain}
\newtheorem{theorem}{Theorem}[section]

\theoremstyle{definition}

\newtheorem{assumption}[theorem]{Assumption}
\theoremstyle{remark}
\newtheorem{remark}[theorem]{Remark}

\usepackage[textsize=tiny]{todonotes}
\renewcommand{\paragraph}[1]{\textbf{#1} \quad}
\newcommand\mydots{\hbox to 1em{.\hss.\hss.}}

\icmltitlerunning{The Role of Entropy and Reconstruction in Multi-View Self-Supervised Learning}

\begin{document}

\twocolumn[
\icmltitle{The Role of Entropy and Reconstruction in\\ Multi-View Self-Supervised Learning }

\icmlsetsymbol{equal}{*}

\begin{icmlauthorlist}
\icmlauthor{Borja Rodr\'iguez-G\'alvez}{kth,a}
\icmlauthor{Arno Blaas}{a}
\icmlauthor{Pau Rodr\'iguez}{a}
\icmlauthor{Adam Goli{\'n}ski}{a}
\\
\icmlauthor{Xavier Suau}{a}
\icmlauthor{Jason Ramapuram}{a}
\icmlauthor{Dan Busbridge}{a}
\icmlauthor{Luca Zappella}{a}
\end{icmlauthorlist}

\icmlaffiliation{kth}{Division of Information Science and Engineering (ISE), KTH
Royal Institute of Technology, Stockholm, Sweden}
\icmlaffiliation{a}{Apple}

\icmlcorrespondingauthor{Borja Rodr\'iguez-G\'alvez}{borjarg@kth.se}
\icmlcorrespondingauthor{Luca Zappella}{lzappella@apple.com}

\icmlkeywords{Machine Learning, ICML}

\vskip 0.3in
]

\printAffiliationsAndNotice{} 

\begin{abstract}

The mechanisms behind the success of multi-view self-supervised learning (MVSSL) are not yet fully understood. 
Contrastive MVSSL methods have been studied through the lens of InfoNCE, a lower bound of the Mutual Information (MI). 
However, the relation between other MVSSL methods and MI remains unclear. 
We consider a different lower bound on the MI %
consisting of an entropy and a reconstruction term (ER), and analyze the main MVSSL families through its lens. 
Through this ER bound, we show that clustering-based methods such as DeepCluster and SwAV maximize the MI. 
We also re-interpret the mechanisms of distillation-based approaches such as BYOL and DINO, showing that they explicitly maximize the reconstruction term and implicitly encourage a stable entropy, and we confirm this empirically. 
We show that replacing the objectives of common MVSSL methods with this ER bound achieves competitive performance, while making them stable when training with smaller batch sizes or smaller exponential moving average (EMA) coefficients. 

Github repo: \href{https://github.com/apple/ml-entropy-reconstruction}{apple/ml-entropy-reconstruction}. %

\end{abstract}

\section{Introduction}
\label{sec:introduction}

Representation learning tackles the problem of learning lower dimensional representations of data which capture the data's semantic information. 
To achieve this, many representation learning methods aim to maximize the \emph{mutual information} (MI) between the input data and the learned representations~\citep{linsker1988self,
belghazi2018mutual,
hjelm2018learning}, while inducing biases in the model that steer the learned information to be semantically meaningful~\citep{
alemi2017deep, 
oord2018representation, 
velickovic2019deep}. 
As such, 
MI has played a crucial role in understanding the performance of many representation learning methods~\citep{tishby2000information, rodriguez2020convex, goldfeld2020information}.

Recently, 
multi-view self-supervised learning (MVSSL),
where the loss enforces the model to produce similar representations for different views of the same data,
has proven to be a successful approach for representation 
learning~\citep{
bachman2019learning,
tian2020contrastive,
he2020momentum,
caron2021emerging}.
The success of MVSSL has motivated the research of several
families of MVSSL approaches, such as \emph{contrastive}~\citep{chen2020simple},
\emph{clustering}-~\citep{caron2018deep}, 
and 
\emph{distillation}-based methods~\citep{grill2020bootstrap}. However, 
 the effort to understand all of them under a common umbrella lags behind the development of new methods.
In this work, 
we aim to further our understanding of %
MVSSL methods 
by identifying %
any mechanisms contributing to maximizing MI, and to what extent they do so.

The connection of the contrastive MVSSL methods to MI maximization is well established through the \infonce bound~\citep{oord2018representation,poole2019variational}, which, in the MVSSL context, lower bounds the MI between the learned representations of different views.
\citet{tian2020makes} and~\citet{tsai2020self} argue that maximizing this MI is attractive as a representation learning target since, when the views are selected carefully, it extracts task-relevant and discards task-irrelevant information. 

The interest in the MI perspective on representation learning, 
and MVSSL in particular, 
has been undermined following the work of \citet{tschannen2019mutual},
whose key result is showing that maximizing MI alone is not sufficient for learning good representations.
Yet, it is empirically evident that methods based on MI lower bound maximization are competitive with state-of-the-art, 
and \citet{tschannen2019mutual} note that
``the performance of these methods depends strongly on the bias that is encoded not only in the encoders, but also on the actual form of the used MI estimators''.
In our opinion, their results strongly motivates further study of the mechanisms by which, 
and to what extent, 
the MI maximization takes place in representation learning.

In this work, we center our analysis of MVSSL methods around the MI between the learned representations of different views $Z_1, Z_2$.
The MI lower bound we focus on consists of an {\em entropy} and a {\em reconstruction} term~\citep{gallager1968information}:
 \begin{align}
     \mi(Z_1;Z_2) \geq
     \underbrace{\ent(Z_2)}_{\textnormal{Entropy}} 
     + \underbrace{\bE
     [\log \sfq_{Z_2|Z_1}(Z_2)]}_{\textnormal{Reconstruction term}}
     \coloneqq \mi_{\er}(Z_1;Z_2),
     \nonumber
 \end{align}
where the 
$\log \sfq_{Z_2|Z_1}$ corresponds to a choice of a similarity function between representations used in MVSSL, e.g., a cosine similarity.
We refer to this bound as \er, referring to the \emph{Entropy} and \emph{Reconstruction} terms.
Focusing on this bound, rather than the \infonce, allows us to analyze a wide range of MVSSL methods through the lens of MI.

The work closest in spirit to ours is
\citep{wang2020understanding}, which analyzes the contrastive MVSSL methods through the lens of \emph{alignment} and \emph{uniformity}, two metrics which they derive through formulating desiderata for the learned representations.
While their motivation was, in the light of the results of \citet{tschannen2019mutual}, to offer an alternative interpretation of \infonce, different than as a lower bound on MI, we show the metrics they define coincide with a specific instantiation of the \ba MI bound we consider. 
We generalize their results through the use of the \ba bound which allows us to also analyze the clustering- and distillation-based MVSSL methods.

Our contributions in this work are the following:
\begin{itemize}
\addtolength\itemsep{-0.3em}
    \item We review how, and to what extent, the major families of MVSSL methods (contrastive, clustering, and distillation-based) maximize MI via the use of the \ba bound on MI.  
    Specifically,
    we show that the clustering-based methods \swav \citep{caron2020unsupervised} and \deepcluster \citep{caron2018deep} maximize the \ba bound and therefore the MI between representations of different views.
    \item \looseness=-1 We empirically %
    show that simply substituting the loss function and instead optimizing \ba in \simclr \citep{chen2020simple}, %
    \byol \citep{grill2020bootstrap}, and \dino \citep{caron2021emerging} results in %
    similar performance while improving resiliency with respect to training with smaller batch sizes or exponential moving average (EMA) coefficients. This is especially important for distillation methods such as \byol or \dino, as they become resilient to batch size changes without any need for hyperparameter changes or gradient accumulation.
    \item Finally, we show that it is not necessary for distillation methods like \byol to maximize entropy to achieve competitive results, 
    although mechanisms such as the softmax centering in \dino and other related architectural constraints prevent the entropy collapse.
\end{itemize}

\section{Background}
\label{sec:background_and_related_work}
Here, we introduce some notation, the multi-view self-supervised learning setting, and the relevant bounds on MI.

\paragraph{Notation}
$X$ represents a random variable (RV) with probability mass function or density
$\sfp_X$, and $x$ is its realization.
Expectations are denoted as $\bE[f(X)] = \bE_{x \sim \sfp_X}[f(x)]$. 
The conditional density 
for a fixed realization $x$ is denoted as $\sfp_{Y|X=x}$.
The density $\sfq_{Y|X}$ is not the real conditional density of $X$ given $Y$, but an an auxiliary one that serves, e.g., as an optimization target.
The mutual information is denoted as
$\mi(X;Y)$, 
the Shannon and the differential entropy 
are both denoted as $\ent(X)$, and
the Kullback-Leibler divergence between densities $p$ and $q$ is denoted as $\kldiv( \sfp \lVert \sfq )$.
A sub-sequence of elements from $a$ to $b$ in a sequence $x$
is denoted as $\ind{x}{a:b}$, 
and 
all elements except $\ind{x}{i}$ as $\ind{x}{\neq i}$.

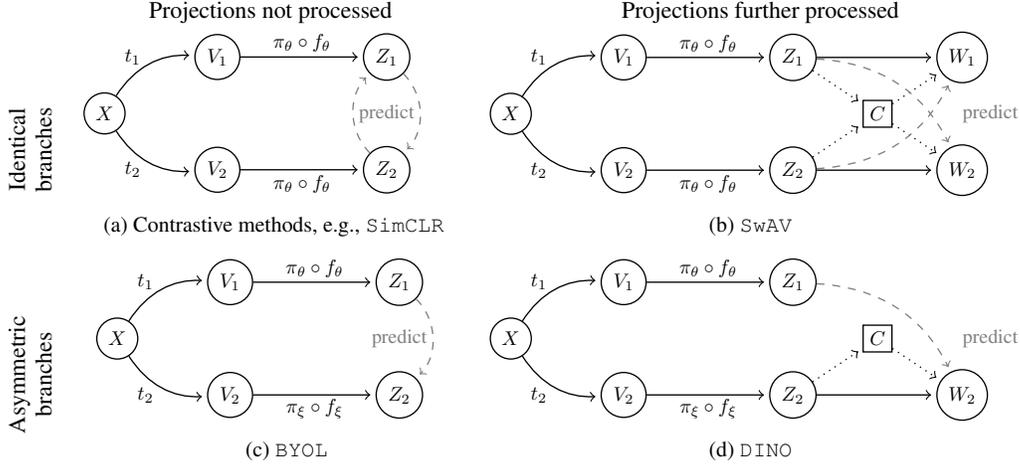
\begin{figure*}[ht]
\centering

\begin{tabular}{ccc}
& \small Projections not processed
& \small Projections further processed
\\
\rotatebox{90}{\parbox{2.5cm}{\centering \small Identical\\branches}}
& \begin{tikzpicture}[->, shorten >=2pt, line width=0.5pt, node distance=2cm, scale=0.75, every node/.style={scale=0.75}]]

    \node[scale=1.1] at (3,-2) {(a) Contrastive methods, e.g., \simclr};
	\node[draw, circle] (X) at (0,0) {$X$};
	\node[draw, circle] (V1) at (2,1) {$V_1$};
	\node[draw, circle] (V2) at (2,-1) {$V_2$};
	\node[draw, circle] (Z1) at (5,1) {$Z_1$};
	\node[draw, circle] (Z2) at (5,-1) {$Z_2$};
	
	\path (X) edge [bend left = 30] (V1);
	\path (X) edge [bend left = -30] (V2);
        \path (V1) edge (Z1);
	\path (V2) edge (Z2);
	\path (Z1) edge [dashed, bend left = 45, color=gray] (Z2);
	\path (Z2) edge [dashed, bend left = 45, color=gray] (Z1);

	\node at (0.5,1) {$t_1$};
	\node at (0.5,-1) {$t_2$};
	\node at (3.5,1.25) {$\pi_\theta \circ f_\theta$};
	\node at (3.5,-1.25) {$\pi_\theta \circ f_\theta$};
 
	\node at (5.0,0) {\textcolor{gray}{{predict}}};
\end{tikzpicture}
& \begin{tikzpicture}[->, shorten >=2pt, line width=0.5pt, node distance=2cm, scale=0.75, every node/.style={scale=0.75}]]

    \node[scale=1.1] at (4.25,-2) {(b) \swav};
	\node[draw, circle] (X) at (0,0) {$X$};
	\node[draw, circle] (V1) at (2,1) {$V_1$};
	\node[draw, circle] (V2) at (2,-1) {$V_2$};
	\node[draw, circle] (Z1) at (5,1) {$Z_1$};
	\node[draw, circle] (Z2) at (5,-1) {$Z_2$};
    \node[draw, circle] (W1) at (8,1) {$W_1$};
	\node[draw, circle] (W2) at (8,-1) {$W_2$};
    \node[draw, rectangle] (C) at (6.5,0) {$C$};
	
	\path (X) edge [bend left = 30] (V1);
	\path (X) edge [bend left = -30] (V2);
        \path (V1) edge (Z1);
	\path (V2) edge (Z2);
    \path (Z1) edge (W1);
	\path (Z2) edge (W2);
    \path (Z1) edge [dotted] (C);
    \path (Z2) edge [dotted] (C);
    \path (C) edge [dotted] (W1);
    \path (C) edge [dotted] (W2);
	\path (Z1) edge [dashed, bend left = 30, color=gray] (W2);
	\path (Z2) edge [dashed, bend left = -30, color=gray] (W1);
	
	\node at (0.5,1) {$t_1$};
	\node at (0.5,-1) {$t_2$};
	\node at (3.5,1.25) {$\pi_\theta \circ f_\theta$};
	\node at (3.5,-1.25) {$\pi_\theta \circ f_\theta$};
	\node at (8.5,0) {\textcolor{gray}{{predict}}};
\end{tikzpicture} 
\\
\rotatebox{90}{\parbox{2.5cm}{\centering \small Asymmetric\\branches}}
& \begin{tikzpicture}[->, shorten >=2pt, line width=0.5pt, node distance=2cm, scale=0.75, every node/.style={scale=0.75}]]

    \node[scale=1.1] at (3,-2) {(c) \byol};
	\node[draw, circle] (X) at (0,0) {$X$};
	\node[draw, circle] (V1) at (2,1) {$V_1$};
	\node[draw, circle] (V2) at (2,-1) {$V_2$};
	\node[draw, circle] (Z1) at (5,1) {$Z_1$};
	\node[draw, circle] (Z2) at (5,-1) {$Z_2$};
	
	\path (X) edge [bend left = 30] (V1);
	\path (X) edge [bend left = -30] (V2);
        \path (V1) edge (Z1);
	\path (V2) edge (Z2);
	\path (Z1) edge [dashed, bend left = 45, color=gray] (Z2);
	
	\node at (0.5,1) {$t_1$};
	\node at (0.5,-1) {$t_2$};
	\node at (3.5,1.25) {$\pi_\theta \circ f_\theta$};
	\node at (3.5,-1.25) {$\pi_\xi \circ f_\xi$};
	\node at (5.0,0) {\textcolor{gray}{{predict}}};
\end{tikzpicture} 
& \begin{tikzpicture}[->, shorten >=2pt, line width=0.5pt, node distance=2cm, scale=0.75, every node/.style={scale=0.75}]]

    \node[scale=1.1] at (4.25,-2) {(d) \dino};
	\node[draw, circle] (X) at (0,0) {$X$};
	\node[draw, circle] (V1) at (2,1) {$V_1$};
	\node[draw, circle] (V2) at (2,-1) {$V_2$};
	\node[draw, circle] (Z1) at (5,1) {$Z_1$};
	\node[draw, circle] (Z2) at (5,-1) {$Z_2$};
	\node[draw, circle] (W2) at (8,-1) {$W_2$};
    \node[draw, rectangle] (C) at (6.5,0) {$C$};
	
	\path (X) edge [bend left = 30] (V1);
	\path (X) edge [bend left = -30] (V2);
        \path (V1) edge (Z1);
	\path (V2) edge (Z2);
	\path (Z2) edge (W2);
    \path (Z2) edge [dotted] (C);
    \path (C) edge [dotted] (W2);
	\path (Z1) edge [dashed, bend left = 30, color=gray] (W2);
	
	\node at (0.5,1) {$t_1$};
	\node at (0.5,-1) {$t_2$};
	\node at (3.5,1.25) {$\pi_\theta \circ f_\theta$};
	\node at (3.5,-1.25) {$\pi_\xi \circ f_\xi$};
	\node at (8.5,0) {\textcolor{gray}{{predict}}};
\end{tikzpicture}
\end{tabular}

\vspace{-5pt}
\caption{\emph{The MVSSL prototypes.} 
An image $X$ is transformed with augmentations $t$ to generate two views $V$ and projections $Z$. 
Dashed and dotted lines indicate loss functions and optional relationships between variables respectively.
\textbf{Top:}
Identical branches: Parameters $\theta$ are identical across branches and the loss is symmetric.
\textbf{Bottom:}
Asymmetric branches: Parameters $\theta,\xi$ across branches are different and the loss is asymmetric.
\textbf{Left:}
The projections $Z$ are not further processed.
\textbf{Right:}
The projections $Z$ 
are processed into auxiliary discrete variables $W$, potentially using another variable $C$. 
Parameters $\theta,\xi$ are optimized such that $Z$ are predictive of the other branch's $W$. 
}

\label{fig:mvssl}
\vspace{-0.5cm}
\end{figure*}

\paragraph{Multi-view self-supervised learning}
In MVSSL, for each data sample $X$, we generate two (or more) views $V_b$. These views are  
commonly obtained by using augmentations \citep{bachman2019learning,tian2020makes,chen2020simple,caron2020unsupervised,zbontar2021barlow}, by leveraging multiple 
modalities~\citep{radford2021learning}, or 
natural views of data~\citep{tian2020contrastive},
e.g., multiple camera views of the same scene.
Views $V_b$ are chosen or engineered such that most of the semantic information remains unchanged 
with respect to 
the original data sample $X$ and shared between the views~\citep{tian2020makes}.
Each view is then passed through a neural network encoder $f_\theta(\cdot)$ to produce representations $R_b$ which are in turn projected via $\pi_\theta(\cdot)$,
usually a small MLP, 
into a lower dimensional space to yield $Z_b$, where $\theta$ are the learnable parameters.
Typically, the intermediate representations $R_b$ are used for downstream tasks and transfer learning, as that yields better performance than using $Z_b$ \citep{chen2020simple,bordes2022guillotine}.
The parameters $\theta$ are learned by optimizing an objective which encourages the projections $Z_b$ to be predictive of the other branches' outputs $Z_{(\neq b)}$. 
This is commonly achieved by optimizing a \emph{similarity} score, such as the L2 distance.
Most of the methods use two views and we will focus on this setting, 
without loss of generality.\footnote{
When more than two views are considered, the objective decomposes into a sum of independent sub-objectives based on view pairs, see e.g.,~\citet{tian2020contrastive} or \citet{caron2018deep}. 
} 
Since the processing of each view takes place separately and for some methods differs between views, we refer to those separate computation paths as \emph{branches}.
See~\Cref{fig:mvssl} for an illustrative diagram.

The three families of MVSSL considered in this work are 
\emph{contrastive}, \emph{clustering-} and \emph{distillation}-based methods.
Contrastive methods work by comparing the projections of the two views of the same %
datum (or \emph{positive pairs}), with a set of projections of different %
data (or \emph{negative pairs}). The different methods in this category are usually distinguished by how they define the negative pairs.
Most of these methods are derived %
either from
the metric learning literature~\citep{sohn2016improved} %
or the
\infonce objective \citep{oord2018representation}, which is a lower bound on the mutual information between the projections $\mi(Z_1;Z_2)$. 
We discuss these methods in detail in \Cref{subsec:contrastive_methods}.
Clustering methods 
cluster the projections %
from one branch and use the resulting discrete cluster assignments as targets for the other branch by optimizing a cross-entropy loss \citep{caron2018deep, caron2020unsupervised, asano2019self}.
Distillation-based methods design the two branches asymmetrically, using one branch's projections as targets for the other %
\citep{grill2020bootstrap,chen2021exploring,caron2021emerging}.
The two branches, referred to as \emph{teacher} and \emph{student}, %
differ. 
Common differences include
gradients being computed only by the student (stop-grad),
teacher's parameters being set via an %
EMA of the student's, 
and an additional predictor network for the student.

\paragraph{Mutual information lower bounds}
Estimating MI is fundamentally difficult~\citep{mcallester2020formal} and for gradient-based representation learning, it is common to rely on the gradients of a lower bound on MI without estimating MI directly \cite{poole2019variational}.
In this work, the core quantity of interest is the MI between MVSSL projections $\mi(Z_1; Z_2)$.
Two MI lower bounds that can be used to optimize this quantity are \infonce 
and \ba.

\looseness=-1 \infonce 
 \cite{oord2018representation,poole2019variational} is a lower bound on MI.
In MVSSL, the MI is between the projections $Z_1, Z_2$.
It is estimated from  a sequence of i.i.d.\ samples of pairs $(\ind{Z_1}{1:k}, \ind{Z_2}{1:k})$ from the joint density $\sfp_{Z_1,Z_2}$:
\begin{equation}
    \mi_{\nce}(Z_1;Z_2) \!\coloneqq\! \frac{1}{k} \sum_{i=1}^k \bE
    \!
    \left[ \log 
    \frac{e^{f(\ind{Z_1}{i},\ind{Z_2}{i})}}{ \frac{1}{k} \sum_{j=1}^k e^{f(\ind{Z_1}{i}, \ind{Z_2}{j})}}
    \right],
    \label{eq:infonce}
\end{equation}

where $f(\cdot,\cdot)$ is a function scoring similarity between vectors, e.g., cosine similarity.
Many contrastive methods use it as a loss function in the original or slightly different forms depending on negative sample choice.
We discuss the MI maximization in this class of methods in detail in Section~\ref{subsec:contrastive_methods}.

The \ba bound is a long standing result in information theory~\citep{gallager1968information}.
It can be derived by considering a
tractable
\emph{reconstruction density} 
$\sfq_{Z_2|Z_1}$ that for MVSSL corresponds to a choice of a similarity function:
\vspace{-3pt}
 \begin{align}
     \!\!
     \mi(Z_1;\!Z_2) 
     \! &= \!
     \bE \!\! \left[ \log 
     \frac{\sfq_{Z_2|Z_1}(Z_2)}{\sfp_{Z_2}(Z_2)} 
     \right] 
     \!\! + \!
     \bE[ \overbrace{\kldiv( \sfp_{Z_2|Z_1} \lVert \sfq_{Z_2|Z_1} )}^{\ge 0}] 
     \nonumber \\
     \! &\geq \! 
     \ent(Z_2)
     \!+\! 
     \bE[\log \sfq_{Z_2|Z_1}\!(Z_2)]
     \!\coloneqq\! \mi_{\er}(Z_1;\!Z_2).
     \label{eq:base_decomp}
 \end{align}
In the MVSSL setting,
$\sfq_{Z_2|Z_1}$ is a design choice and we are interested in optimizing the parameters of $\pi_\theta \circ f_\theta$ such that the resulting density $\sfp_{Z_1,Z_2}$ maximizes $\mi_{\ba}(Z_1;Z_2)$.
The density $\sfp_{Z_1,Z_2}$ implicitly results from sampling inputs $X$, possibly transforming them via stochastic transformations $t$, and then deterministically transforming them through the encoder $\pi_\theta \circ f_\theta$ to form $Z$.
The term $\bE[\kldiv( \sfp_{Z_2|Z_1} \lVert \sfq_{Z_2|Z_1} )]$ determines the magnitude of the gap of the $\mi_{\ba}$ bound.

\looseness=-1 The term \emph{reconstruction} originates from information theory. 
It is often concerned with reconstructing a signal from a compressed code and is equal to $-\ent(Z_2|\hat{Z_2})$, where $\hat{Z_2}$ is a RV such that $Z_2 - Z_1 - \hat{Z_2}$ is a Markov chain. 
We find it also more appropriate to reason about MVSSL such as the right column of~\autoref{fig:mvssl}, 
where $Z_1$ and $W_2$ belong to different spaces, and hence the term \emph{similarity} seems less accurate.

Intuitively, the \emph{entropy} and \emph{reconstruction} terms in the \er bound~\eqref{eq:base_decomp} play different roles in MVSSL. The entropy term determines how much information from one projection \emph{can be learnt}, while the reconstruction term determines how much of this available information \emph{is learnt}. For instance, let the projections lay on the sphere: the more spread out (higher entropy) the projections of different data are, the more revealing (higher mutual information) it is if projections from different views of the same datum are close (lower reconstruction error). Conversely, 
if one branch projects all data to the same point 
(lowest entropy, also known as \emph{collapse}), 
the projections from the other branch can’t reveal any information about them. 

\paragraph{MVSSL for small batch sizes} \looseness=-1 Small batch sizes degrade the performance of MVSSL methods, especially contrastive ones~\citep{chen2020simple, grill2020bootstrap, caron2021emerging}. Potentially, this is due to the fact that most methods maximize the entropy either explicitly or implicitly, as shown in this paper, and the entropy estimation is limited to $\log k$ bits for a batch size of $k$~\citep{mcallester2020formal}. Some works \citep{haochen2021provable, chen2021simpler, yuan2022provable} addressed this issue and modified existing methods to perform well under the small batch size regime.

\section{MVSSL and MI optimization}
\label{sec:mvssl_and_mi_optimization}

In this section, we reflect on the relationship between different MVSSL methods and the MI. First, we review the known connection between contrastive methods and MI maximization through the \infonce bound, as well as the lack thereof. Also, we show that none of the existing methods formally maximize the \ba bound, while all of them are a good proxy for it. 
Next, we show for the first time that the clustering-based methods \deepcluster~\citep{caron2018deep} and \swav~\citep{caron2020unsupervised} also optimize the MI through the \ba bound. 
Finally, we interpret the techniques used in distillation-based methods such as %
EMA~\citep{grill2020bootstrap} and softmax centering~\citep{caron2021emerging} as mechanisms to prevent the entropy collapse. The results of this section are summarized in~\Cref{tab:mi}.

\subsection{Contrastive methods}
\label{subsec:contrastive_methods}

Contrastive learning (CL) methods are the family of MVSSL methods that have been most closely connected to MI maximization in the existing literature and, as such, a good starting point for our analysis. Here, we first give a review of the connections established through the \infonce bound and otherwise, before exhibiting the relationship to the \ba bound. %
Summarizing, generally CL algorithms cannot be formally shown to maximize the \infonce nor the \ba bound due to the violation of the i.i.d. assumption. This is not the case for \cmc those methods derived from it, nor for methods using a memory bank like Instance Discrimination~\citep[\ir]{wu2018unsupervised} or \moco~\citep{he2020momentum, chen2020improved} under particular circumstances, which do maximize the \infonce. Nevertheless, as also concluded by~\citet{wang2020understanding}, CL is a good proxy for entropy maximization, and therefore, for MI maximization.

Given the %
projection of a view of %
datum $i$, e.g., $\ind{Z_2}{i}$, contrastive learning algorithms aim to maximize its similarity with the projection of another view of the same %
datum, e.g., $\ind{Z_1}{i}$ ({\em positive sample}), while making it as different as possible from the %
projections of a set of \emph{negative samples} $\cN(\ind{Z_2}{i})$. This is achieved by minimizing a cross entropy loss based on a similarity score. Given a batch of $k$ samples a generic contrastive loss for the second branch is
\begin{equation}
    \cL_{\textnormal{contr,2}} \coloneqq - \frac{1}{k} \sum_{i=1}^k \log  \frac{e^{f(\ind{Z_2}{i},\ind{Z_1}{i})}}{ \sum_{Z' \in \cN(\ind{Z_2}{i})} e^{f(\ind{Z_2}{i},Z')}} 
    \label{eq:contrastive_loss}
\end{equation}
and the full loss is $\cL_{\textnormal{contr}} \coloneqq (\cL_{\textnormal{contr,1}} + \cL_{\textnormal{contr,2}})/2$, where usually $f = \csim(\cdot) / \tau$, $\csim(\cdot)$ is the cosine similarity, and $\tau$ is a temperature parameter. Then, different CL methods are distinguished by how the set of negative samples for a particular sample $\ind{Z_2} {i}$ is constructed.
Note that the negatives  might include samples from the other branches.

In \cmc~\citep{tian2020contrastive}, the negative samples set is composed of all the other projections from the opposite branch, i.e., $\cN(\ind{Z_2}{i}) = \ind{Z_1}{1:k}$. %
Comparing~\eqref{eq:infonce} and~\eqref{eq:contrastive_loss} with these negative samples we see that \cmc maximizes the \infonce bound and $\bE[-\lcmc] \leq \mi(Z_1;Z_2) - \log k$.

The maximization of the \infonce bound can be similarly shown for methods that can be derived from the basic \cmc, like the full \cmc, where more than two views are considered; \citep{bachman2019learning}, which adapts \dimm~\citep{hjelm2018learning} to the basic \cmc; and~\citep{tian2020makes}, which attempts to learn the augmentations that best suit the information maximization.

For \simclr~\citep{chen2020simple}, on the other hand, the negative samples are all the projections other than $\ind{Z_2}{i}$, i.e., $\cN(\ind{Z_2}{i}) = \ind{Z_2}{\neq i} \cup \ind{Z_1}{1:k}$. %
Given such a definition of the negative set, even if all negative samples were identically distributed, the negative samples are not independent as $\ind{Z_1}{j}$ and $\ind{Z_2}{j}$ are derived from the same datum $j$, for all $j$s. 
As shown in~\citep{tschannen2019mutual}, \infonce is not maximized when violating the independence assumption. %
Hence, \simclr does not maximize the \infonce bound.
This also holds true for methods that are derived from \simclr such as \citep{ramapuram2021stochastic}.

Finally, methods like \ir or \moco use representations from a memory bank as negative samples, i.e., $\cN(\ind{Z_2}{i}) = \ind{Z_\textnormal{bank}}{1:m}$. In these cases the negative samples can be dependent and are not identically distributed with respect to $\ind{Z_2}{i}$. However, \citet{wu2020conditional} showed that under certain mild conditions on the distribution of these samples the contrastive loss used in these methods is a lower bound on the \infonce, and thus optimizing it also maximizes MI.

\paragraph{Relationship with the \ba bound}
None of the contrastive methods above directly translates to an optimization of the \ba bound, even if it may appear so. 
In the context of \eqref{eq:contrastive_loss}, 
if we consider a density s.t.\ $\sfq_{Z_2|Z_1=z_1}(z_2) \propto \exp {f(z_2,z_1)}$, the expected value of the first term corresponds to the reconstruction error in~\eqref{eq:base_decomp}, and when $f(\cdot,\cdot)$ is the cosine similarity with temperature $\tau$, the density $\sfq_{Z_2|Z_1=z_1}$ corresponds to a von Mises--Fisher density with mean direction $z_1$ and concentration parameter $1/\tau$. 
However, as shown above, in all methods analyzed, the negative samples are either not independent between themselves (as in \simclr), or not identically distributed with respect to the positive sample (as in \moco), or the set contains the positive pair itself (as in \cmc). 
Therefore, the 
log-denominator in~\eqref{eq:contrastive_loss} 
is not an unbiased kernel density estimator (KDE, ~\citet{joe1989estimation}) of the entropy and therefore its expectation is not necessarily the entropy $\ent(Z_2)$ from~\eqref{eq:base_decomp}.

Nonetheless, all these methods %
force the projections to be maximally separated from the negative samples in a convex set (usually the hypersphere). 
Moreover, the highest entropy distribution on a convex set is precisely the uniform distribution on that volume. 
Hence, the contrastive loss, even with non-i.i.d. negative samples, is a good proxy for entropy maximization, and therefore, for MI maximization. 
\citet{wang2020understanding} make a similar observation and conclude that maximizing the uniformity of the samples in the projections’ space is required for good performance.

\paragraph{Caveats} As seen above, most current analyses for CL methods require the i.i.d. assumption, which is not usually met due to the use of batch normalization. The breaking of the independence assumption is important as it can break the \infonce results~\citep{tschannen2019mutual, wu2020conditional}. Nonetheless, it does not discredit that the result of the KDE is a good proxy to maximize the entropy.

\subsection{Clustering-based methods}
\label{susec:clustering_methods}

In this section, we show that both \deepcluster~\citep{caron2018deep, asano2019self} and \swav~\citep{caron2020unsupervised} maximize the \er lower bound on the MI between the projections of different views of the data $\mi_{\er}(Z_1;Z_2)$. 

The key observation underlying the results in this section is that \deepcluster and \swav generate a discrete surrogate of the projections, e.g., for the second branch $W_2 = \phi(Z_2)$, and that they maximize the \ba bound on $\mi(Z_1;W_2) \leq \mi(Z_1;Z_2)$, where the inequality holds by the data processing inequality. For the rest of the section, let $\cZ \subseteq \bR^d$ and $\cW = \{ 1, \ldots, m \}$.

\deepcluster has an asymmetric setting with $\xi = \theta$ (\Cref{fig:mvssl}d). First, the cluster assignments $\ind{W_2}{i} = \phi(\ind{Z_2}{i})$ of all the $n$ data points are obtained solving the problem

\begin{equation*}
    C^\star \in \arginf_{C \in \bR^{d \times m}} \frac{1}{n} \sum_{i=1}^n \lVert \ind{Z_2}{i} - C \ind{\sfp_2}{i} \rVert^2,
\end{equation*}
with $\ind{\sfp_2}{i} \in \{0, 1\}^m$ and $\lVert \ind{\sfp_2}{i} \rVert_0 = 1$, where $C^\star$ represent the $m$ centroids of the clusters in $\cZ$ and $\ind{\sfp_2}{i}$ is the p.m.f. of $\ind{W_2}{i}$ given $\ind{Z_2}{i}$.\footnote{\citet{asano2019self} obtain the clusters solving an optimal transport problem similar to \swav.} Then, the parameters $\theta$ are optimized by minimizing the cross entropy
\begin{equation*}
    \ldeepcluster \coloneqq - \frac{1}{k} \sum_{i=1}^k \Big( \ind{\sfp_2}{i} \Big)^\intercal \log \Big( \sfs \circ g_\theta ( \ind{Z_1}{i})  \Big),
\end{equation*}
where $g_\theta: \cZ \to \bR^m$ is a small predictor network, and $\sfs$ is the softmax operator.
Note that $Z$ also depends on $\theta$ via $Z 
\!= \!\pi_{\theta} \!\circ\! f_{\theta} (V)$, see~\Cref{fig:mvssl}.
With $\sfq_{W_2|Z_1=z_1} = \sfs \circ g_\theta(z_1)$, \emph{this optimization precisely amounts to maximizing the reconstruction term in the \ba bound for $I(Z_1;W_2)$}.  
Furthermore, to prevent degenerate solutions, \citet{caron2018deep} sample the images of each batch based on a uniform distribution over cluster assignments, i.e. for each batch $\sfp_{W_2} \approx \frac{1}{k} \sum_{i=1}^k \sfp_2^{(i)}$ is almost uniform. Through this, \emph{the entropy $\ent(W_2)$ is approximately maximized}. Combined with the maximization of the reconstruction term via $\ldeepcluster$, %
this implies \emph{\deepcluster maximizes the \ba MI bound}.

Now, let us turn to \swav.
\swav has a symmetric setting (\Cref{fig:mvssl}b). We focus on branch $b=2$, as the analysis is analogous for the other branch.  Here, the cluster assignments $\ind{W_2}{i} = \phi(\ind{Z_2}{i})$ are obtained solving the following optimization problem
\begin{equation*}
    P_2 = \argmax_{P \in \cP} \bigg\{ \textnormal{Tr} \Big(\ind{Z_2}{1:k} C^\intercal P^\intercal \Big) + \epsilon \ent(P) \bigg\},
\end{equation*}
where $\ind{Z_2}{1:k} \in \bR^{k \times d}$, $C \in \bR^{m \times d}$ are the $m$ centroids (or prototypes) in $\bR^d$, $\cP = \{ P \in \bR_+^{k \times m}:  P^\intercal \bm{1}_k = \bm{1}_m / m \textnormal{ and } P \bm{1}_m = \bm{1}_k / k \}$ is the transportation polytope, and $\bm{1}_k$ is the all ones vector in $\bR^k$. Let $\ind{C}{i}$ and $\ind{P_2}{i}$ denote the $i$-th row of $C$ and $P_2$, respectively. In \swav, both the projections and the prototypes lay in the unit hypersphere, i.e., $\ind{Z}{i}, \ind{C}{i} \in \bS^{d-1}$, and thus maximizing the dot product is equivalent to minimizing the squared $\ell_2$ norm distance ~\citep{grill2020bootstrap}. %
Moreover, to aid the optimization calculations, an entropic regularization is included to approximately solve it using the Sinkhorn-Knopp algorithm~\citep{sinkhorn1974diagonal,cuturi2013sinkhorn},
where $\ent(P_2) \coloneqq - \sum_{i=1}^k \Big(\ind{P_2}{i}\Big)^\intercal \log \ind{P_2}{i}$.

The $l$-th element of $\ind{P_2}{i}$ can be understood as the probability of assigning $\ind{Z_2}{i}$ to the cluster $\ind{W_2}{i} = l$. The optimization aims to have $P_2 \in \cP$ and therefore $P_2^\intercal \bm{1}_k \approx \bm{1}_m / m$, which by this interpretation would mean that 
$\sfp_{W_2} \approx \bm{1}_m /m$ is approximately uniform, thus maximizing the entropy $\ent(W_2)$. Therefore,
this construction \emph{maximizes the desired entropy $\ent(W_2)$ in the \er bound}

For \swav, similarly to \deepcluster, \emph{the reconstruction term is maximized} by minimizing the loss function
\begin{equation*}
    \lswavtwo \coloneqq - \frac{1}{k} \sum_{i=1}^k \Big(\ind{\sfp_2}{i}\Big)^\intercal \log \Big( \sfs \big( C \ind{Z_{1}}{i} \big) \Big),
\end{equation*}
where $\ind{\sfp_2}{i} = \ind{P_2}{i} / (\bm{1}_m^\intercal \ind{P_2}{i})$  and $\sfq_{W_2|Z_1=z_1} = \sfs(Cz_1)$, hence maximizing the mutual information $\mi(Z_1;W_2)$. 
An analogous analysis for the branch $b=1$ reveals that minimizing $\lswavone$ with the entropic regularisation assignment maximizes the mutual information $\mi(W_1;Z_2)$. 
In \swav, 
the prototypes are treated as parameters of the network (i.e., $C \in \theta$) and are updated using stochastic gradient descent to minimize $\lswav$.
This implies \emph{\swav also maximizes \er}.

\subsection{Distillation methods}
\label{subsec:distillation_methods}

Distillation methods naturally optimize the reconstruction term of the \ba bound since the projection of one branch is optimized to predict the projection of the other branch. However, it is more challenging to understand if and how they might maximize the entropy term of \ba, hence, we cannot yet claim they are maximizing the MI. There are some tools, such as EMA or centering, that distillation methods employ that could have an effect on the entropy. In fact, such tools are key to prevent the phenomenon known as collapse~\citep{grill2020bootstrap, caron2021emerging}. %
Our analysis of their role below does not yield definitive, formal statements. However, it should still shed some light on this question.

First, let us detail how each method maximizes the reconstruction term of the \ba bound.
We start by analyzing the reconstruction term for the \byol loss, which is the $\ell_2$ normalised mean squared error
\begin{equation}
    \lbyol \coloneqq \frac{1}{k} \sum_{i=1}^k \Big \lVert \overline{g_\theta(\ind{Z_1}{i})} - \overline{\ind{Z_2}{i}} \Big \rVert^2,
    \label{eq:byol}
\end{equation}
where $\overline{x} \coloneqq  x / \lVert x \rVert$. Since $\lVert \overline{x} - \overline{y} \rVert^2 = 2(1-\csim(x,y))$, optimizing~\eqref{eq:byol} is equivalent to maximizing the reconstruction term in the \ba bound with a von Mises--Fisher reconstruction density with mean direction $\overline{g_\theta(\ind{Z_1}{i})}$ and concentration parameter 1.
For \dino, the loss is similar to the one used by the clustering-based methods, namely
\begin{equation}
    \ldino \coloneqq - \frac{1}{k} \sum_{i=1}^k \sfs \big( (\ind{Z_2}{i} - C) / \tau_2 \big)^\intercal \log \Big( \sfs (\ind{Z_1}{i} / \tau_1 ) \Big),
    \label{eq:dino}
\end{equation}
where $C$ is a centering variable, and $\tau_1, \tau_2$ are temperature hyperparameters. Letting $\sfp_{W_2|Z_2 = z_2} = \sfs \big( (z_2 - C) / \tau_2 \big)$ and $\sfq_{W_2|Z_1=z_1} = \sfs(z_1/ \tau_1)$ shows that optimizing~\eqref{eq:dino} is equivalent to maximizing the reconstruction term in the \ba bound of $\mi(Z_1;W_2) \leq \mi(Z_1;Z_2)$.

Let us now analyze the potential effect of the stabilizing algorithms used by distillation methods on the entropy of the projections to understand if distillation methods also maximize the entropy term of the \ba bound. We focus on the role of EMA and centering.

EMA introduces an asymmetry between the teacher and the student in distillation methods (\Cref{fig:mvssl}b and d). Specifically, the teacher's parameters $\xi$ track the student's parameters $\theta$ during the optimization with the use of EMA: $\xi \leftarrow \lambda \xi + (1 - \lambda) \theta$ for some $\lambda \in (0,1)$ close to 1. The hypothesis is two-fold: on the one hand, while $\xi$ does depend on $\theta$, the dependence is weak enough so that $\ent(Z_2)$ or $\ent(W_2)$ is not degrading to values yielding trivial bounds. This would happen in the extreme case of $\xi = \theta$, for which minimizing the respective losses will have an optimal solution $\theta^\star$ that would be highly concentrated or degenerate around one point, under which $\ent(Z_2) \to - \infty$ or $\ent(W_2) = 0$, which clearly would not maximize the MI.
On the other hand, the dependence of $\xi$ on $\theta$, while weak, ensures that the projections $Z_2$ capture information about the data.
If this was not the case, e.g., by fixing $\xi$ to random values, the then random projections $Z_2$ would contain very little information about $X$. In this case, despite maximising $\mi(Z_1;Z_2)$ via minimising the respective losses and simultaneously ensuring constant entropy $\ent(Z_2)$ (due to the random projections), the information learned would still be little as by the data processing inequality $\mi(Z_1; Z_2) \leq \mi(X;Z_2)$.
\byol and \dino balance this trade-off between not maximizing MI due to minimal entropy and maximizing MI to a small achievable minimum with constant entropy with their choice of $\lambda$, but the resulting effect on entropy and MI maximization is hard to estimate.

Beyond EMA, \dino also promotes a high conditional entropy $\ent(W_2|Z_2)$ through the centering before the softmax operation. Like in \swav, this avoids collapse as it controls the entropy $\ent(W_2)$ via $\ent(W_2|Z_2) \leq \ent(W_2)$. To be precise, the center $C$ in~\eqref{eq:dino} is updated with an EMA of the previous projections, that is, $C \leftarrow \mu C + \frac{1 - \mu}{k} \sum_{i=1}^k \ind{Z_2}{i}$ for some $\mu \in (0,1)$. Then, the right balance between this EMA and the temperature parameters $\tau_1$ and $\tau_2$ adjusts how uniform the conditional density $\sfp_{W_2|Z_2}$ is. This promotes a high conditional entropy $\ent(W_2|Z_2)$. However, having a completely uniform conditional density means that $\sfp_{W_2|Z_2} = \sfp_{W_2}$ and thus no information of $Z_2$ is in $W_2$. For this reason, \citet{caron2021emerging} need to also include a sharpening of the conditional density via the temperature $\tau_2$. Therefore, the degree of maximization of $\ent(W_2)$ is hard to quantify as it depends on the chosen values of the parameters $\mu, \tau_1,$ and $\tau_2$.

To summarize, the use of both EMA and centering is crucial for distillation methods to work, and they do affect the entropy term of the \ba bound. However, it is not yet possible to quantify these effects exactly, hence, one cannot make any statement that distillation methods maximize MI, despite clearly maximizing the reconstruction term of the \ba bound.

\begin{table}[t]
\centering
\caption{\emph{The 
relation between existing MVSSL methods and the maximization of MI via the \infonce and \er lower bounds.}\\
\checkmark: formally shown,
(\checkmark): approximately or empirically,
$\times$: no formal or empirical evidence, 
$^{*}$: previously known (\Cref{subsec:contrastive_methods}).}
\label{tab:mi}
\small
\begin{tabular}{lccc}
\toprule
 Model & \infonce & \er & Violation \\
\midrule
\cmc & \checkmark${}^*$ & (\checkmark) & -\\
\simclr & $\times$ & (\checkmark) & negatives not i.i.d. \\
\ir, \moco & (\checkmark)${}^*$ & (\checkmark) & negatives not i.i.d.\\
\midrule
\deepcluster & $\times$ & \checkmark & -\\
\swav& $\times$ & \checkmark & -\\
\midrule
\byol & $\times$ & (\checkmark) & not max. entropy\\
\dino & $\times$ & (\checkmark) & not max. entropy\\
\bottomrule
\end{tabular}
\vspace{-0.3cm}
\end{table}

\section{Optimizing the \ba bound in practice}
\label{sec:optimizing_ba_bound}

\looseness=-1 In this section, we describe different ways to maximize the \ba bound regardless of the MVSSL prototype (see~\Cref{fig:mvssl}). That is, we will describe how to estimate the entropy and the reconstruction term in~\eqref{eq:base_decomp} when the projections are not processed (\Cref{fig:mvssl}a and c). The case when discrete surrogates are generated (\Cref{fig:mvssl}b and d) is discussed in \Cref{app:mse_convergence_discrete}. Then, the objective resulting from such an estimation is maximized. Later, in~\Cref{sec:experiments}, we use these approaches on top of the architectures of current contrastive and distillation-based methods and observe that their performance is on par (or slightly better) than their original formulation, and that they become more resilient to the choice of the batch size and EMA coefficient without the need for neither adjusted hyper-parameters nor accumulated gradients.

\subsection{Maximizing MI between projections}
\label{subsec:mi_between_projections}

We consider an estimation of the \ba bound of the MI between the projections $\mi_\er(Z_1;Z_2)$. 
Let $f(z_2,z_1)$ be a function measuring the similarity between $z_1$ and $z_2$. Choosing the reconstruction density  %
$\sfq_{Z_2|Z_1=z_1}(z_2) \propto \exp f(z_2,z_1)$, an unbiased estimate of the reconstruction term is given by
\begin{equation}
    \widehat{\textnormal{Rec}}_\textnormal{cont} \coloneqq \frac{1}{k} \sum\nolimits_{i=1}^k f(\ind{Z_2}{i}, \ind{Z_1}{i}),  \label{eq:estimate_reconstruction_cont}
\end{equation}
where the term associated with the normalizing constant of the density is discarded as it does not affect the optimization.
To estimate the entropy term, one may consider different variants of KDEs. 
For example, both the KDE of~\citet{joe1989estimation}
\begin{equation} 
    \label{eq:entropy_estimate}
    \hat{\ent}(Z_2)_{\textnormal{KDE,Joe}} \coloneqq - \frac{1}{k} \sum_{i=1}^k \log \hat{\sfp}_{Z_2}(\ind{Z_2}{i})
\end{equation}
or the plug-in estimator~\citep{krishnamurthy2015lectures}  
\begin{equation}
    \label{eq:ent_plug_in_cont}
    \hat{\ent}(Z_2)_{\textnormal{KDE,plug-in}} \coloneqq -  \sum_{i=1}^k \hat{\sfp}_{Z_2}(\ind{Z_2}{i}) \log \hat{\sfp}_{Z_2}(\ind{Z_2}{i})
\end{equation}
can be used (both give similar results in practice, see \Cref{app:differnt_continuous_estimators}).
Here, $\hat{\sfp}_{Z_2}(z)$ is \citet{joe1989estimation}'s KDE of $p_{Z_2}$:
\begin{equation}
    \label{eq:kde_density}
    \hat{\sfp}_{Z_2}(z) \coloneqq \frac{1}{k h^d} \sum_{j=1}^k \sfq \bigg( \frac{z - \ind{Z_2}{j}}{h} \bigg),
\end{equation}
with kernel $\sfq(\cdot)$ and bandwidth $h \in \bR_+$.
\looseness=-1 Both the reconstruction and the entropy estimators are (asymptotically) unbiased and converge in mean squared error (MSE) with an appropriate choice of the bandwidth (see~\Cref{app:convergence}). The selection of an optimal kernel bandwidth can be seen as a limitation of \er. While minimizing the number of hyper-parameters would be desirable, the bandwidth plays a similar role to the temperature term typically tuned in other SSL methods, e.g.~\citep{chen2020simple}. So much so, that we adopted as bandwidth the same temperature parameter specified by the SSL methods on top of which we incorporate \er.

\paragraph{Connection to CL}
When the chosen kernel $\sfq$ is such that $\sfq(z_2 - z_1) = f(z_2,z_1)$, then maximizing the \ba bound with estimators
(\ref{eq:estimate_reconstruction_cont},\,\ref{eq:entropy_estimate})
is \emph{equivalent to contrastive learning} with the negative samples being $\cN(\ind{Z_2}{i}) = \ind{Z_2}{\neq i}$, up to constants independent of the optimization parameters.

\paragraph{Connection to Uniformity and Alignment} 
The \emph{alignment} and \emph{uniformity} objective of~\citet{wang2020understanding} is a relaxation of the \ba objective with estimators~(\ref{eq:estimate_reconstruction_cont},\,\ref{eq:entropy_estimate}). Let $f(z_2, z_1) = \lVert z_2 - z_1 \rVert_2^\alpha$, then the estimator~\eqref{eq:estimate_reconstruction_cont} recovers their alignment term. 
Consider also a kernel $\sfq(z_2-z_1) \propto \exp \big( - t \lVert z_2 - z_1 \rVert_2^2 \big)$, then~\citet{joe1989estimation}'s KDE~\eqref{eq:entropy_estimate} recovers their alignment term after applying Jensen's inequality.\footnote{The application of Jensen's inequality makes \citet{wang2020understanding}'s objective a looser MI lower bound than the \ba bound.} Hence, our analysis can be considered a natural extension of their analysis to other MVSSL families.

\paragraph{Connection to Identifiability}
Under certain assumptions, MVSSL partitions the latent representations into a content component, invariant to augmentations, and a style component, which can change with augmentations~\citep{von2021self}. The \ba objective recovers their main theorem (Theorem~4.4) with a reconstruction density $\sfq_{Z_2|Z_1=z_1}(z_2) \propto \exp \big({- \lVert z_2 - z_1 \rVert_2^2}\big)$. Moreover, CL methods implicitly invert the underlying generative model of the observed data, again under certain assumptions \citep{zimmermann2021contrastive}. We show that the same is true for methods maximising the \ba bound, revealing that the main reason for this inversion is not the contrastive nature of the methods, but that they maximize the mutual information (see \Cref{app:theoretical_properties}).

\vspace{-0.2cm}
\subsection{Dealing with an EMA}
\label{subsec:dealing_with_ema}

The maximization of the \ba bound is compatible with an asymmetric structure (\Cref{fig:mvssl}c, d) where the teacher's parameters $\xi$ are updated with an EMA of the student's parameters $\theta$. 
The objective is equivalent to the maximization of the symmetric bound with an additional \texttt{stop\_gradient} operator on the teacher's projections. 
The optimization from the reconstruction of the teacher from the student is unaffected. Then, since the entropy of the student's projections $Z$ (or surrogates $W$) is maximized, it will also be maximized for the teacher, which is only updated through the EMA. This is confirmed empirically in~\Cref{sec:experiments}.

\section{Experiments}
\label{sec:experiments}
In this section, we show that replacing the objective of common MVSSL methods with the \ba bound results in competitive performance while being more robust to the changes in batch size and EMA coefficient without changing any other hyperparameters. Further experiments are included in \Cref{app:extended_results,app:different_weights} and the code is available at \href{https://github.com/apple/ml-entropy-reconstruction}{https://github.com/apple/ml-entropy-reconstruction}.

\paragraph{Experimental Setup}
For all experiments, we pre-train a \texttt{resnet50}~\citep{he2016deep} on the ImageNet~\citep{deng2009imagenet} training set. We train for 400 epochs and following~\citet{chen2020improved} we use a batch size of 4096 with the \texttt{LARS} optimizer~\citep{you2017large} with linear warmup, a single cycle cosine annealed learning rate schedule, and a base learning rate of $0.3$ ~\citep{goyal2017accurate} . 
We chose \byol, \dino, and \simclr as baseline methods, with \cmc results presented in \Cref{app:extended_results}. For each model except \dino, we substitute their objective function by the continuous estimate of the \ba bound from~\Cref{sec:optimizing_ba_bound},\footnote{We use the plug-in estimator instead of~\citet{joe1989estimation}'s, but we observe both to perform almost identically (\Cref{app:differnt_continuous_estimators}).} while keeping the original set of augmentations and their original projection heads. For \dino we estimate the entropy as the average of the discrete plug-in entropy among replicas. \cmc shares  augmentations and projection head with \simclr. %

\begin{table}[t]
\centering
\caption{\textit{Training with \ba yields competitive performance while improving stability with small batch size and EMA coefficients.} \textbf{Model}: set of augmentations, loss, and projection head. $^*$Our implementation. \textbf{\er}: the original loss has been substituted by the \ba bound~\eqref{eq:base_decomp}. \textbf{MI}: known to maximize MI. ($\checkmark$): no formal proof~(\Cref{subsec:dealing_with_ema}). $\mathbf{\Delta 512}$: accuracy drop with batch size 512. \textbf{$\Delta$EMA$_{0.8}$}: accuracy drop with EMA coefficient of $0.8$.}
\label{tab:imnet_acc}
\small
\begin{tabular}{lccccc}
\toprule
        Model &             MI &  Acc ($\uparrow$) &  $\Delta 512 (\downarrow)$ & $\Delta \text{EMA}_{0.8} (\downarrow)$ \\
\midrule
        \dino &              ? &             75.59 &                       6.76 &                                   8.25 \\
  \dino + \er & $(\checkmark)$ &             73.39 &                       2.35 &                                   0.92 \\\hline
        \byol &              ? &             73.42 &                      23.65 &                                   2.63 \\
  \byol + \er & $(\checkmark)$ &             71.94 &                       2.35 &                                   0.41 \\\hline
      \simclr &       $\times$ &             70.23 &                       2.17 &                                      - \\
\simclr + \er &   $\checkmark$ &             70.86 &                       1.01 &                                      - \\
\bottomrule
\end{tabular}
\vspace{-0.4cm}
\end{table}

\paragraph{Training with \er yields competitive accuracy}  We train a linear classifier on top of the ImageNet pre-trained features and report the test accuracy in \Cref{tab:imnet_acc}. For all models, we kept their original hyperparameters.  For \simclr, adding \ba increases %
test accuracy ($+0.72$) while for \byol and \dino it %
decreases slightly ($-1.5$ and $-1.65$, respectively).

\begin{figure*}[t]
    \centering
    \includegraphics[width=0.8\linewidth,trim={0 0 0 0},clip]{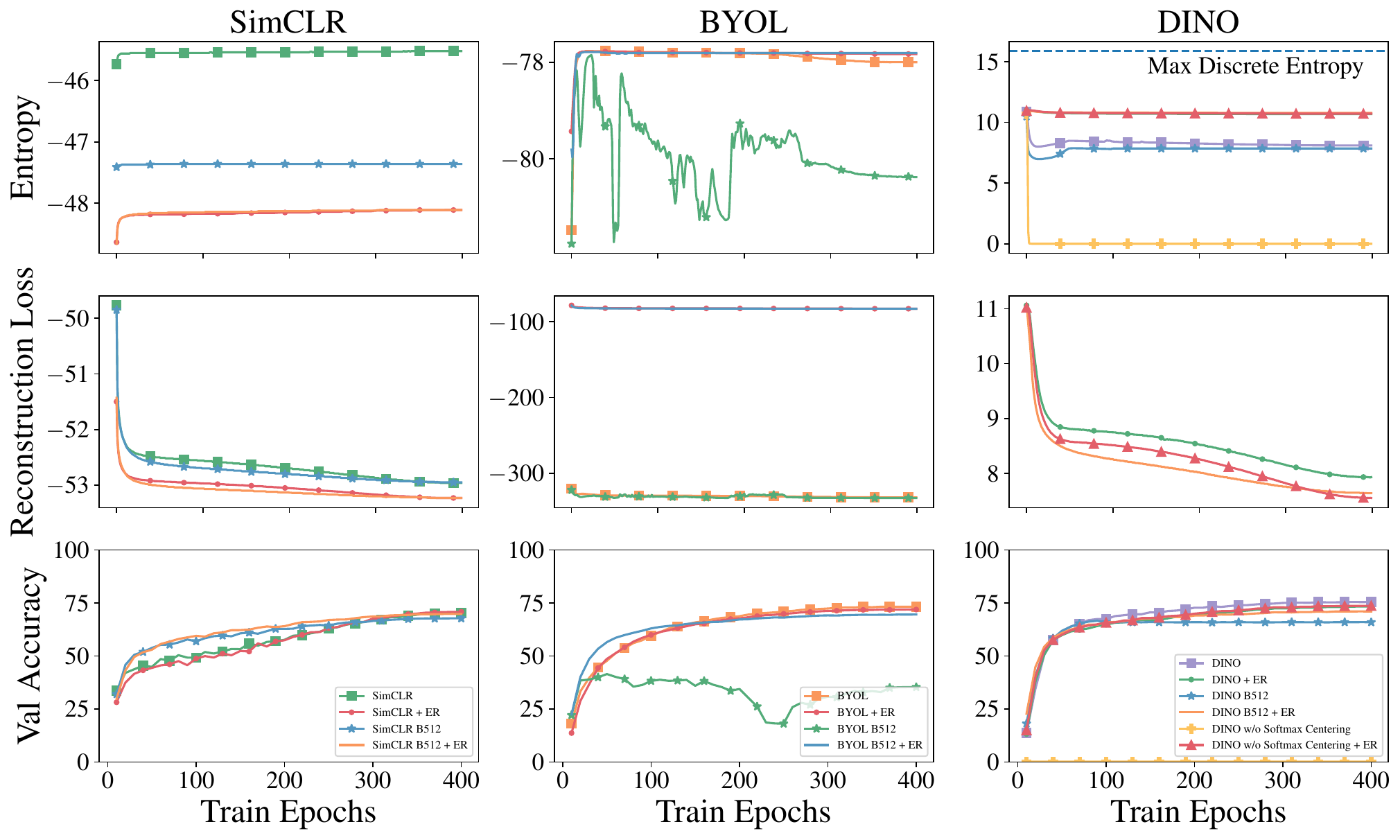}
    \caption{\textit{\ba maximizes entropy during training (top) while it is unclear for distillation methods. \ba allows training \dino w/o softmax centering.} Top: Entropy dynamics while training \simclr, \byol, \dino w/ and w/o \ba, and \dino w/ and w/o softmax centering for 400 epochs. Middle: Reconstruction loss dynamics. Bottom: top-1 accuracy on the ImageNet test set (linear probe trained online).}
    \label{fig:entropy_methods_no_discrete}
\vspace{-0.4cm}
\end{figure*}

\paragraph{\ba further improves distillation method's stability with small batch size and small EMA coefficients} The right column in \Cref{tab:imnet_acc} shows the performance degradation when training with batch size $512$ and EMA coefficient of $0.8$ instead of $0.99$ (we observe similar results with a batch size 1024 or an EMA coefficient of $0.6$). The original version of \byol and \dino exhibit the largest degradation of all algorithms. This can also be observed in \Cref{fig:entropy_methods_no_discrete}. Note that \citet{grill2020bootstrap} provided recipes to train \byol with smaller batch sizes by retuning hyperparemeters or by gradient accumulation. They also observed that the batch size had a strong influence on the optimal EMA coefficient. Here, we limit our observation to what happens when nothing else is changed in the optimization. Interestingly, we observe that \er significantly improves the resilience towards the change in batch size for all methods tested, especially for \byol where the degradation is reduced from $-20.32$ to $-0.21$. Regarding the EMA coefficient, we observe a degradation of $-8.25$ for \dino and $-2.62$ for \byol which are reduced to $-0.92$ and $-0.41$ respectively with \er.

In fact, we find that training with \er outperforms recent literature on small-batch SSL training \citep{haochen2021provable,chen2021simpler,yuan2022provable}. For example, for \simclr with batch size 512, we report an accuracy of $69.85$ (\Cref{tab:imnet_acc}) while the most recent of these works reports an accuracy of $68.8$~\citep{yuan2022provable}. 

\paragraph{BYOL does not maximize entropy} \Cref{fig:entropy_methods_no_discrete} shows the evolution of entropy and reconstruction during training (top and middle) and the ImageNet accuracy (bottom) (see \Cref{app:discrete_mvssl} for clustering methods like \deepcluster and \swav). We observe that methods trained with \ba clearly maximize entropy while others such as \byol with batch size 4096 display a slight decrease in entropy while still achieving high accuracy. This might provide an empirical answer to the question left in~\Cref{subsec:distillation_methods} and indicate that \byol does not maximize entropy.  
The EMA was introduced to avoid representation collapse in the absence of negative samples.
When properly tuned, the effect seems sufficient to maintain a high entropy and create discriminative representations. Nevertheless, one could argue that it does not take full advantage of the overall space (or we would observe higher entropy) and that the accuracy is very sensitive to its tunning (see \Cref{tab:imnet_acc} and \Cref{fig:entropy_methods_no_discrete}).
In addition to the EMA, \dino introduces a softmax centering procedure to keep the output probabilities in a certain range. In \Cref{fig:entropy_methods_no_discrete}, we observe that \dino's entropy and accuracy become extremely low when softmax centering is deactivated. Notably, \textit{adding \er makes it possible to train \dino without softmax centering}, which confirms that softmax centering plays a role in keeping the entropy high (\Cref{subsec:distillation_methods}).

\paragraph{ER is not sensitive to the entropy estimator} All \er models except \dino  used a KDE-based entropy estimator. To gain more insight into the effect of the %
estimator, we train a continuous KDE-based version of \dino + \er and compare it with the one reported in \Cref{tab:imnet_acc}, which uses an exact discrete estimator. We find no significant differences %
between their performances (see \Cref{app:extended_results}).

\section{Discussion}
\label{sec:discussion}
We showed to what extent different MVSSL methods maximize MI through the \er bound on the MI.
First, we revisited previous knowledge about the maximization of MI in contrastive methods and reinterpreted it in the context of \er.
Second, we showed that two clustering-based methods, \deepcluster and \swav, maximize the \er bound.
Third, we interpreted two distillation-based methods, \byol and \dino, as maintaining a stable level of entropy while maximizing the reconstruction term of the \er bound.

\looseness=-1 We explained how \er can be optimized in most MVSLL frameworks, and we showed empirically that \simclr, \byol and \dino, when optimizing the \er bound result in a performance which is competitive with that of the respective original versions. We also showed that it is not necessary for distillation methods like \byol to maximize entropy to achieve competitive results. 
This is an interesting observation in the context of \citep{wang2020understanding} who conclude both alignment and uniformity are required for contrastive methods to work well, we showed that at least for distillation methods, maximizing uniformity is not necessary. Uniformity (or high entropy), however, seems to be correlated with resilience as all methods became more resilient to smaller batch size and/or EMA coefficient when maximizing \er, with a particularly pronounced effect for distillation methods. Understanding the exact mechanism for these behaviors remains an exciting subject of future work.

Finally, our theoretical analysis in \Cref{subsec:mi_between_projections} and \Cref{app:theoretical_properties} indicates that methods that explicitly maximize the \ba bound should yield desirable identifiability properties. We believe that exploring this result in practice is an exciting avenue for future research.

\section*{Acknowldegments}

The authors thank the reviewers for their valuable feedback, which resulted in new experiments and clarifications that strengthened 
the paper,
as well as
the colleagues at Apple for productive discussions that helped shape and fortify the paper, especially  Effrosyni Simou, Michal Klein, Tatiana Likhomanenko, and R. Devon Hjelm.

Borja Rodríguez-Gálvez was funded, in part, by the Swedish research council under contract 2019-03606.

\bibliography{bibliography.bib}
\bibliographystyle{icml2023}

\newpage
\appendix
\Large
\textbf{Appendices}
\normalsize

\section{Entropy and reconstruction estimators MSE convergence}
\label{app:convergence}
In this section, we describe the MSE behaviour of the entropy and reconstruction estimators of~\Cref{sec:optimizing_ba_bound}.

\subsection{Estimators on a ``continuous space"}
\label{app:mse_convergence_continuous}

\subsubsection{Entropy estimation and selection of the bandwidth parameter}
\label{app:entropy_estimation_cont}

The bias and variance of \citet{joe1989estimation}'s KDE estimator $\hat{\ent}_{\textnormal{KDE,Joe}}$ of the entropy $\ent(Z_2)$ are~\citep[Section 4, page 695]{joe1989estimation}
\begin{align*}
    \bB[\hat{\ent}_{\textnormal{KDE,Joe}}] &\in \cO(k^{-1}h^{4-d}) + \cO(k^{-2}h^{-2d}) + \cO(h^4) \textnormal{ and} \\
    \bV[\hat{\ent}_{\textnormal{KDE, Joe}}] &\in \cO(k^{-1}) + \cO(k^{-2} h^{8-d}) + \cO(k^{-2}h^{-d}) \\
    &\quad + \cO(k^{-1}h^{8-d}) + \cO(k^{-2} h^{4-2d}) + \cO(h^8).
\end{align*}

Hence, as long as $h \in \cO(k^{-1/(d+\varepsilon)})$ for some small $\varepsilon > 0$ both the bias and the variance vanish, and the estimator convergences in MSE, even if it does so at a slow rate. Then, a sensible choice of the bandwidth is $h \approx 1$ since $k^{-1/(d+\varepsilon)} \to 1$ as $d$ increases.

Under the mild assumption that the distribution of $Z_b$ is $\beta$-smooth (i.e., it belongs to the H\"older or Sobolev classes) then the bias and variance of both KDE estimators $\hat{\ent}_{\textnormal{KDE}}$ are~\citep{krishnamurthy2015lectures}
\begin{align*}
    \bB[\hat{\ent}_{\textnormal{KDE}}] &\in \cO(h^\beta) \textnormal{ and} \\
    \bV[\hat{\ent}_{\textnormal{KDE}}] &\in \cO(k^{-1} h^{-d}).
\end{align*}

As previously, the bias and the variance of the estimator only vanish if $h \in \cO(k^{-1/(d+\varepsilon)})$ for some small $\varepsilon > 0$, with the optimal choice $h = k^{-1/(d+2\beta)}$. Nonetheless, having a bias term independent of the parameter of the optimisation is not harmful in itself. Hence, when the KDE estimator is employed only for optimisation purposes both $h \in \cO(k^{-1/(d+\varepsilon)})$ and $h \in \cO(1)$ may work. For instance, for the experiments using the von Mises--Fisher distribution we set $h = 0.1$ to match the temperature employed by \citep[\cmc]{tian2020contrastive} and \citep[\simclr]{chen2020simple}.

\subsubsection{Reconstruction estimation}
\label{app:reconstruction_estimation_cont}

Note that $\log \sfq_{Z_2 | \ind{Z}{i}_{1}}(\ind{Z}{i}_2)$ are independent and identically distributed random variables with expectation $\bE[\log \sfq_{Z_2 | Z_{1}}(Z_2)]$. Hence, the empirical estimator is unbiased. Similarly, the variance of the estimator is $\bV[\frac{1}{k} \sum_{i=1}^k \log \sfq_{Z_2 | \ind{Z}{i}_{1}}(\ind{Z}{i}_2)] = \sigma^2_\sfq / k $, where the individual variance is $\sigma^2_\sfq =\bV[\log \sfq_{Z_2 | Z_{1}}(Z_2)]$.

Consider now that a reconstruction density is of the form $\sfq_{Z_2|Z_{1} = z_{1}}(z_2) = C e^{-\rho(z_2,z_{1})}$ and that the projections lay in a convex body $\cZ \in \bR^d$. Then, we know that $\log \sfq_{Z_2|Z_{1} = z_{2}}(z_1) \in [\log C - \rho(\cZ), \log C]$, where $\rho(\cZ)$ is the diameter of $\cZ$ with respect to $\rho$. Therefore, by the Popoviciu's inequality on variances we have that $\sigma^2_\sfq \leq \rho(\cZ)^2 / 4$, which implies that for $\rho(\cZ) < \infty$ the estimator converges in MSE. This holds for the two cases considered in this paper:
\begin{itemize}
    \item Von Mises--Fisher distribution in $\cZ = \bS^{d-1}$: Here the diameter with respect to $\rho(z_1,z_2) = \kappa \csim(z_1,z_2)$ is $\rho(\cZ) = \kappa^2$ and hence the estimator converges in MSE at a $\kappa^2/(4k)$ rate.
    \item Gaussian distribution in $\cZ = [-1,1]^d$: Here the diameter with respect to $\rho(z_1,z_2) = \lVert z_1 - z_2 \rVert^2 / (2 \sigma^2)$ is $\rho(\cZ) = 2d/\sigma^2$ and hence the estimator converges in MSE at a $d/(2k\sigma^2)$ rate.
\end{itemize}

\begin{remark}
    Consider, without loss of generality from the setting in \Cref{sec:optimizing_ba_bound}, a reconstruction density of the form $\sfq_{Z_2|Z_{1} = z_{1}}(z_2) = C(z_1) e^{-\rho(z_2,z_{1})}$. Then, to be precise, the reconstruction estimate in~\eqref{eq:estimate_reconstruction_cont} is biased with bias $\bE[\log C(Z_1)]$. However, since this term does not affect the optimization, we say the estimator is unbiased, meaning that is unbiased to the terms required for the optimization.

    To see why the term does not affect the optimization, assume first that $Z_1$ is fixed to $z_1$. Then, $C(z_1)$ is a constant and clearly optimizing $\bE[f(Z_2,Z_1)|Z_1=z_1] + \log C(z_1)$ is equivalent to optimizing just $\bE[f(Z_2,Z_1)]$. Then, note that $\bE[f(Z_2,Z_1) + \log C(Z_1)] = \bE_{z_1 \sim p_{Z_1}} [ \bE[f(Z_2, Z_1)|Z_1=z_1] + \log C(z_1)]$, and hence optimizing $\bE[f(Z_2,Z_1)]$ is sufficient.
\end{remark}

\subsection{Estimators in a discrete space}
\label{app:mse_convergence_discrete}

\subsubsection{Maximizing MI between projections and surrogates}
\label{subsec:mi_between_projections_and_surrogates}

Now, we consider an estimation of the \ba bound for the MI between the projections and the discrete surrogates $\mi(Z_1;W_2)$. We will assume that the discrete surrogates are in $[d]$. As previously, we may consider the following unbiased estimate of the reconstruction term
\begin{equation}
     \widehat{\textnormal{Rec}}_\textnormal{disc} \coloneqq \frac{1}{k} \sum_{i=1}^k \log \sfq_{W_2|Z_1=\ind{Z_1}{i}}(\ind{W_2}{i}),
    \label{eq:estimate_reconstruction_dist}
\end{equation}
where the reconstruction density could simply be, for instance, $\sfq_{W_2|Z_1=z} = \sfs(z)$.
Since $W_2$ is a discrete RV, one may consider the empirical estimate of the marginal, i.e., $\hat{\sfp}_{W_2} \coloneqq \frac{1}{k} \sum_{i=1}^k \sfp_{W_2 | Z_2 = \ind{Z_2}{i}}$, and use the unbiased plug-in estimate~\citep{girsanov1959property}  of the entropy:
\begin{equation}
    \label{eq:ent_plug_in_disc}
    \hat{\ent}(W_2)_{\textnormal{plug-in}} \coloneqq - \frac{1}{k} \sum_{i=1}^k \hat{\sfp}_{W_2}(\ind{W_2}{i}) \log \hat{\sfp}_{W_2}(\ind{W_2}{i}).
\end{equation}

Both the reconstruction and entropy estimators are unbiased and converge in MSE, see the following sections.

\begin{remark}
    In practice, if the product of the batch size $k$ and the discrete dimension $d$ is large, one may instead consider an average of plug-in estimates. For simplicity, assume that the batch size is a multiple of $r \in \bN$, then this estimate is
    \begin{equation}
        \label{eq:ent_plug_in_avg_disc}
        \hat{\ent}(W_2)_{\textnormal{plug-in-avg}} \coloneqq \frac{1}{r} \sum_{j=1}^r \hat{\ent}(W_2)_{\textnormal{plug-in}}^{(j)}
    \end{equation}
    where $\hat{\ent}(W_2)_{\textnormal{plug-in}}^{(j)}$ is the plug-in entropy estimate of the $j$-th chunk of the batch size data. Usually, each entropy estimation is done in a different machine, or \emph{replica}.
\end{remark}

\subsubsection{Entropy estimation}
\label{app:entropy_estimation_disc}

The plug-in estimator $\hat{\ent}_{\textnormal{PI}}$ of the entropy $\ent(W_2)$ is known to have the following bias and variance terms (see e.g.~\citep[Equations~(3) and (4)]{girsanov1959property} or~\citep[Introduction]{antos2001convergence}):
\begin{align*}
    \bB[\hat{\ent}_{\textnormal{plug-in}}] &\in  \cO\Big( \frac{d-1}{2k} \Big) + \cO \Big( \frac{1}{k^2} \Big) \textnormal{ and}\\
    \bV[\hat{\ent}_{\textnormal{plug-in}}] &\in  \cO\Big( \frac{\sigma^2_\sfp}{k} \Big) + \cO \Big( \frac{1}{k^2} \Big),
\end{align*}
where $\sigma^2_\sfp = \bV[-\log \sfp_{W_2}(W_2)]$. The bias and the variance vanish as long as $d$ is fixed and $\sigma^2_\sfp < \infty$, meaning that the estimator converges in MSE.

Note that $\sfp_{W_2} = \bE[\sfs(Z_2)]$, where $\sfs$ is the softmax operator. Hence, we have that
\begin{align*}
    \bV[-\log \sfp_{W_2}(W_2)] &\leq \bE[\log^2 \sfp_{W_2}(W_2)] \\
    &\leq \bE \bigg[ \Big( Z_{2,W_2} - \log \Big( \sum_{l=1}^d e^{Z_{2,l}} \Big) \Big)^2 \bigg] \\
    &\leq \bE \big[ (\log d + Z_{2,\textnormal{max}} - Z_{2,\textnormal{min}} )^2 \big],
\end{align*}
\looseness=-1 where the first inequality follows from the fact that $\bV[X] \leq \bE[X^2]$; the second from Jensen's inequality and the formula of the softmax; and the last one from the log-sum-exp trick. Here, $Z_{2,l}$ denotes the $l$-th element of the random vector $Z_2$.

In the particular case where the projections lie in the sphere $\bS^{d-1}$ we have that $\sigma^2_p \leq (\log d + 1)^2$. Similarly, if they lay in the cube $[-1, 1]^d$, we have that $\sigma^2_p \leq (\log d + 2)^2$. Therefore, under these standard conditions the variance vanishes at a rate in $\cO(\log^2(d) / k) + \cO(1/k^2)$.

\begin{remark}
    If the average of plug-in estimators is used instead, we may note that this estimator will have a drop in bias and variance proportional to the number of replicas $r$, where the variance is not quadratically affected due to the variance reduction effect of the averaging. More precisely,
    \begin{align*}
        \bB[\hat{\ent}_{\textnormal{plug-in-avg}}] &\in  \cO\Big( \frac{r(d-1)}{2k} \Big) + \cO \Big( \frac{r}{k^2} \Big) \textnormal{ and}\\
        \bV[\hat{\ent}_{\textnormal{plug-in-avg}}] &\in  \cO\Big( \frac{\sigma^2_\sfp}{k} \Big) + \cO \Big( \frac{r}{k^2} \Big).
    \end{align*}
\end{remark}

\subsubsection{Reconstruction estimation}
\label{app:reconstruction_estimation_disc}

As in~\Cref{app:reconstruction_estimation_cont}, note that $\log \sfq_{W_2 | \ind{Z}{i}_{1}}(\ind{W}{i}_2)$ are independent and identically distributed random variables with expectation $\bE[\log \sfq_{W_2 | Z_{1}}(W_2)]$. Hence, the empirical estimator is unbiased. Similarly, the variance of the estimator is $\bV[\frac{1}{k} \sum_{i=1}^k \log \sfq_{W_2 | \ind{Z}{i}_{1}}(\ind{W}{i}_2)] = \sigma^2_\sfq / k $, where $\sigma^2_\sfq = \bV[\log \sfq_{W_2 | Z_{1}}(W_2)]$. Hence, the variance vanishes as long as $\sigma^2_\sfq < \infty$, meaning that the estimator converges in MSE.

As for the entropy estimation, note that $\sfq_{W_2|Z_{1}} = \sfs(Z_{1})$. Hence, repeating the analysis above in~\Cref{app:entropy_estimation_disc} we obtain that $\sigma^2_\sfq \leq \bE \big[ (\log d + Z_{1,\textnormal{max}} - Z_{1,\textnormal{min}} )^2 \big]$ and therefore for projections in the sphere $\bS^{d-1}$ or the cube $[-1,1]^d$ the variance vanishes at a rate in $\cO(\log^2(d) / k) + \cO(1/k^2)$.

\section{Properties of maximizing MI via BA}
\label{app:theoretical_properties}
This section formalises and contextualises statements in \Cref{sec:optimizing_ba_bound}. 

\subsection{Recovering the true latent variables}
\label{subapp:recoer_latent_variables}

Let us consider the standard assumption in independent components analysis (ICA), namely that the data $X$ is generated by a nonlinear, invertible generative process $X = g(\tilde{Z})$ from some original latent variables $\tilde{Z}$. Assume further that the different views from the image can be understood as $V_1 = g(\tilde{Z}_1)$ and $V_2 = g(\tilde{Z}_2)$, where there is some joint density of the latent variables $\sfp_{\tilde{Z}_1,\tilde{Z}_2}$. The next theorem shows how~\citet{zimmermann2021contrastive} theory can be adapted to prove that mutli-view SSL methods that maximize the mutual information between their projections $\mi(Z_1;Z_2)$ can obtain projections equivalent to the true latent variables up to affine transformations.

\begin{theorem}
    \label{th:recovering_true_latents}
    Assume that the latent variables and the network's projections lay on a convex body $\cZ \in \bR^d$. Further assume that the latent variables' marginal distribution is uniform and that the conditional density is described by a semi-metric $\rho$ as $\sfp_{\tilde{Z}_2|\tilde{Z}_1=\tilde{z}_1}(\tilde{z}_2) = C(\tilde{z}_1) e^{-\rho(\tilde{z}_1,\tilde{z}_2)}$. Now let the reconstruction density match the conditional density up to a constant scaling factor $\sfq_{Z_2|Z_1=z_1}(z_2) = C_h(\tilde{z}_1) e^{- \alpha \rho(z_1,z_2)}$. If the generative process $g$ and the parameterised network functions $\pi \circ f$ are invertible and differentiable, and the parameters $\theta$ maximize the lower bound~\eqref{eq:base_decomp} of the mutual information $\mi(Z_1;Z_2)$, then the projections are equivalent to the true latent variables up to affine transformations.
\end{theorem}

\begin{proof}
As in~\citep{zimmermann2021contrastive}, let $h = \pi \circ f \circ g$ be a helper function that brings the true latent variables to the projections so that $Z_1 = h(\tilde{Z}_1)$ and $Z_2 = h(\tilde{Z}_2)$. 

Disregard for a moment the entropy term $\ent(Z_2)$. From \citep[Proposition~4]{zimmermann2021contrastive} we know that if the reconstruction term is maximized (the cross entropy is minimised) then $\rho(\tilde{z}_1,\tilde{z}_2) = \alpha \rho(h(\tilde{z}_1),h(\tilde{z}_2))$ and $C(\tilde{z}_1) = C_h(\tilde{z}_1)$. Moreover, from \citep[Theorem~4]{zimmermann2021contrastive} we further know that $h$ is an invertible affine transformation; i.e. $Z_2 = A\tilde{Z}_2 + b$ for some $A \in \bR^{d \times d}$ and some $b \in \bR^d$. 

Now note that 
\begin{align*}
    \ent(Z_2) &= - \bE \Big[ \log \bE \Big[ C_h(\tilde{Z}_1) e^{- \alpha \rho(h(\tilde{Z}_1), h(\tilde{Z}_2)} \Big] \Big] \\
    &= - \bE \Big[ \log \bE \Big[ C(\tilde{Z}_1) e^{- \rho(\tilde{Z}_1, \tilde{Z}_2)} \Big] \Big] = \ent(\tilde{Z}_2).
\end{align*}
Then, since the latent variables' are uniformly distributed, their entropy is maximal $\ent(\tilde{Z}) = \log |\cZ|$.

Therefore, the unique family of maximizers of the reconstruction term recover the latent variables up to affine transformations are maximizers of the entropy, and hence are the unique family of maximizers of the mutual information. Indeed, take some other maximizer of the entropy, if it is not from this family, it is not a maximizer of the reconstruction and therefore the resulting mutual information is lower. 
\end{proof}

\begin{remark}
    Following the same reasoning and supporting on~\citet{zimmermann2021contrastive}'s theory, we may note that in the particular case that the semi-metric $\rho$ is described by an $L^p$ norm, then the projections are equivalent to the true latent variables up to generalised permutations; that is, $Z = A \tilde{Z}$ for some $A \in \bR^{d \times d}$ such that $(Az)_i = \alpha \beta_i z_{\sigma(i)}$, where $\alpha \in \bR$, $\beta_i \in \{1,-1\}$, and $\sigma$ is a permutation. Similarly, in the more restrictive case that the projections are in the sphere $\cZ = \bS^{d-1}$ and the conditional densities are von Mises--Fisher densities, then the projections are equivalent to the true latent variables up to linear transformations; that is, $Z = A \tilde{Z}$ for some $A \in \bR^{d \times d}$ such that $A^\intercal A = \alpha I$ for some $\alpha \in \bR$.  
\end{remark}

\subsection{Empirical evaluation of identifiability properties}
\label{subapp:experiments}

\begin{table*}[tb]
    \centering
    \caption{Identifiability up to affine transformations, replicated from Table 1 of \citet{zimmermann2021contrastive}
    and updated with results trained using the \ba objective, rather than \infonce.
    The numbers for the \infonce column are taken from \citet{zimmermann2021contrastive} with the exception of the last two rows marked with $^*$, where we were not able to replicate their results.
    The \ba configurations for $h=1$ marked with $^\dagger$ exhibited training instability which is why their results are considerably lower.
    }
    \resizebox{0.8\textwidth}{!}{%
    \begin{tabular}{cccccccc}
        \toprule
        \multicolumn{3}{c}{Generative process $g$} & \multicolumn{2}{c}{Model $f$} & \multicolumn{3}{c}{$R^2$ Score [\%]} \\
        Space & $p(\cdot)$ & $p(\cdot|\cdot)$ & Space & $q_h(\cdot|\cdot)$ & \infonce & $\ba_{h=10}$ & $\ba_{h=1}$ \\
        \midrule
        Sphere & Uniform & vMF($\kappa{=}1$) & Sphere & vMF($\kappa{=}1$) & $99.42$ & $89.31$ & $98.94$ \\
        Sphere & Uniform & vMF($\kappa{=}10$) & Sphere & vMF($\kappa{=}1$) & $99.86$ & $99.87$ & $99.80$ \\
        Sphere & Uniform & Laplace($\lambda{=}0.05$) & Sphere & vMF($\kappa{=}1$) & $99.91$ & $99.88$ & $99.62$ \\
        Sphere & Uniform & Normal($\sigma{=}0.05$) & Sphere & vMF($\kappa{=}1$) & $99.86$ & $99.82$ & $99.31$ \\
        \midrule
        Box & Uniform & Normal($\sigma{=}0.05$) & Unbounded & Normal & $99.60$ & $99.53$ & $71.47^\dagger$ \\
        Box & Uniform & Laplace($\lambda{=}0.05$) & Unbounded & Normal & $99.64$ & $99.57$ & $17.30^\dagger$ \\
        Box & Uniform & Laplace($\lambda{=}0.05$) & Unbounded & GenNorm($\beta{=}3$) & $99.70$ & $99.76$ & $99.73$ \\
        Box & Uniform & Normal($\sigma{=}0.05$) & Unbounded & GenNorm($\beta{=}3$) & $99.69$ & $99.72$ & $99.67$ \\
        \midrule
        Sphere & Normal($\sigma{=}1$) & Laplace($\lambda{=}0.05$) & Sphere & vMF($\kappa{=}1$) & $99.02$ & $99.09$ & $98.77$ \\
        Sphere & Normal($\sigma{=}1$) & Normal($\sigma{=}0.05$) & Sphere & vMF($\kappa{=}1$) & $99.02$ & $98.96$ & $98.43$ \\
        \midrule          
        Unbounded & Laplace($\lambda{=}1$) & Normal($\sigma{=}1$) & Unbounded & Normal & $89.85^*$ & $88.86$ & $88.69$ \\ 
        Unbounded & Normal($\sigma{=}1$) & Normal($\sigma{=}1$) & Unbounded & Normal & $95.26^*$ & $89.89$ & $89.74$ \\
        \bottomrule
    \end{tabular}}
    \label{tab:results_linear}
\end{table*}

\begin{table*}[tb]
    \centering
    \caption{Identifiability up to generalized permutations, replicated from Table 2 of \citet{zimmermann2021contrastive} and updated with results trained using the \ba objective rather than \infonce.
    The numbers for the \infonce column are taken from \citet{zimmermann2021contrastive} with the exception of the rows marked with $^*$, where we were not able to replicate their results.
    }
    \resizebox{0.8\textwidth}{!}{%
    \begin{tabular}{cccccccc}
        \toprule \multicolumn{3}{c}{Generative process $g$} & \multicolumn{2}{c}{Model $f$} & \multicolumn{3}{c}{MCC Score [\%]} \\
        Space & $p(\cdot)$ & $p(\cdot|\cdot)$ & Space & $q_h(\cdot|\cdot)$  & \infonce & $\ba_{h=10}$ & $\ba_{h=1}$ \\
        \midrule
        Box & Uniform & Laplace($\lambda{=}0.05$) & Box & Laplace & $98.62$ & $98.49$ & $97.49$ \\
        Box & Uniform & GenNorm($\beta{=}3$; $\lambda{=}0.05$) & Box & GenNorm($\beta{=}3$)  & $99.90$ & $99.90$ & $95.51$ \\
        \midrule
        Box & Uniform & Normal($\sigma{=}0.05$) & Box & Normal  & $99.77$ & $99.74$ & $96.22$ \\
        Box & Uniform & Laplace($\lambda{=}0.05$) & Box & Normal  & $99.76$ & $99.76$ & $99.74$ \\
        Box & Uniform & GenNorm($\beta{=}3$; $\lambda{=}0.05$) & Box & Laplace  & $98.80$ & $98.77$ & $98.65$ \\
        \midrule
        Box & Uniform & Laplace($\lambda{=}0.05$) & Unbounded & Laplace & $98.57$ & $98.57$ & $98.53$ \\
        Box & Uniform & GenNorm($\beta{=}3$; $\lambda{=}0.05$) & Unbounded & GenNorm($\beta{=}3$)  & $60.54^*$ & $61.23$ & $51.44$ \\
        \midrule
        Box & Uniform & Normal($\sigma{=}0.05$) & Unbounded & Normal  & $58.26$ & $56.52$ & $53.14$ \\
        Box & Uniform & Laplace($\lambda{=}0.05$) & Unbounded & Normal  & $59.67$ & $56.32$ & $31.10$ \\
        Box & Uniform & Normal($\sigma{=}0.05$) & Unbounded & GenNorm($\beta{=}3$)  & $54.59^*$ & $54.58$ & $39.01$ \\
        \bottomrule
    \end{tabular}}
    \label{tab:perm_results}
\end{table*}

In order to empirically study the consequences of the theoretical results of the previous section,
we replicate the results of the experiments reported in \citep[Tables 1 and 2]{zimmermann2021contrastive} 
and study the impact of using the \ba objective on the identifiability properties.

The results are reported in \Cref{tab:results_linear,tab:perm_results}.
The results of \citet{zimmermann2021contrastive} are reported in the column \infonce as they train their unsupervised models with the \infonce objective.
In these experiments, we used \citet{joe1989estimation}'s entropy estimator \eqref{eq:entropy_estimate} 
and we set the KDE bandwidth to $h=1$ or $h=10$ as per \eqref{eq:kde_density} (we report both results).

At a high level, our results indicate that using the \ba objective yields qualitatively similar identifiability properties as using the \infonce objective,
as predicted by theory.
One discrepancy of note is the difference for $h=10$ in the first row of \autoref{tab:results_linear} since that is the setting for which the generative process and the model match the assumptions of \autoref{th:recovering_true_latents}.
However, for $h=1$, the performance is close to the once achieved by \infonce.
Other settings, including those which break the theoretical assumptions, do not exhibit significant differences between \ba and \infonce.

\subsection{Isolating semantic from irrelevant information}
\label{subapp:isolating_semantic}

Similarly to~\Cref{subapp:recoer_latent_variables}, let us consider that the data $X$ is generated by a nonlinear, invertible generative process $X = g(\tilde{Z})$ from some original latent variables $\cZ$ and that the different views can be understood as $V_1 = g(\tilde{Z}_1)$ and $V_2 = g(\tilde{Z}_2)$, where there is some joint density of the latent variables $\sfp_{\tilde{Z}_1, \tilde{Z}_2}$.

Assume that the latent variables can be written as $\tilde{Z} = [{S}, {U}]$, where ${S} \in \bR^{d}$ is some semantic (or content) variable, ${U} \in \bR^{d_u}$ is some irrelevant (or style) variable, and $[ \cdot ]$ denotes the concatentation operation. Furthermore, let us adopt the assumptions from~\citet{von2021self} for the content-preserving conditional density $\sfp_{\tilde{Z}_2|\tilde{Z}_1}$.

\begin{assumption}[Content-invariance]
\label{ass:content_invariance}
The conditional density $\sfp_{\tilde{Z}_2|\tilde{Z}_1}$ of the latent variables of different views has the form
\begin{equation*}
    \sfp_{\tilde{Z}_2|\tilde{Z}_1=\tilde{z}_1}(\tilde{z}_2) = \delta({s}_2 - {s}_1) \sfp_{{U}_2|{U}_1={u}_1}({u}_2),
\end{equation*}
for all $\tilde{z}_1 = [{s}_1, {u}_1]$ and $ \tilde{z}_2 = [{s}_2, {u}_2]$ in $\cZ$ and where $\sfp_{{U}_2|{U}_1={u}_1}$ is continuous for all ${u}_1 \in \bR^{d_u}$.
\end{assumption}

\begin{assumption}[Style changes]
\label{ass:style_change}
Let $\cA$ be the set of subsets of irrelevant variables $A \subseteq \{ 1, \ldots, d_u \}$ and let $\sfp_A$ be a density on $\cA$. Then, the conditional density $\sfp_{{U}_2|{U}_1}$ is obtained via sampling $a \sim \sfp_A$ and letting
\begin{align*}
    \sfp_{{U}_2|{U}_1, A = {u}_1, a}({u}_2) = \delta({u}_{2,a^c}, {u}_{1,a^c}) \sfp_{{U}_{2,a}|{U}_{1,a} = {u}_{1,a}} ({u}_{2,a}),
\end{align*}
where $\sfp_{{U}_{2,a}|{U}_{1,a} = {u}_{1,a}}$ is a continuous density for all ${u}_{1,a} \in \bR^{|a|}$, and where ${u}_{2,a}$ is an abuse of notation to refer to the elements of ${u}_2$ indexed by $a$, and analogously for ${u}_1$ and for $a^c$, which is a shortcut for $\cA \setminus a$.
\end{assumption}

Then, the next theorem shows how~\citet{von2021self} can be adapted to prove that multi-view SSL methods that maximize the mutual information between the projections $\mi(Z_1; Z_2)$ can obtain projections that capture and isolate the semantic information of the true latent variables.

\begin{theorem}
    \label{th:recovering_semantic_info}
    Consider \Cref{ass:content_invariance} and \Cref{ass:style_change} and further assume that
    \begin{itemize}
        \item[1.] the generative process $g$ is smooth, invertible and with a smooth inverse (i.e., a diffeomorphism);
        \item[2.]  $\sfp_{\tilde{Z}}$ is a smooth, continuous density on $\cZ$ with $\sfp_{\tilde{Z}} > 0$ a.e.; and 
        \item[3.] for any $j \in \{1, \ldots, n_u \}$, there is an $a \subseteq \{1, \ldots, n_u \}$ such that $j \in a$, $\sfp_A(a) > 0$, $\sfp_{{U}_{2,a}|{U}_{1,a}={u}_{1,a}}({u}_{2,a})$ is smooth with respect to both ${u}_{1,a}$ and ${u}_{2,a}$, and for any ${u}_{1,a}$ it holds that $\sfp_{{U}_{2,a}|{U}_{1,a}={u}_{1,a}}({u}_{2,a}) > 0$ for all ${u}_{2,a}$ in some open, non-empty subset containing ${u}_{1,a}$.
    \end{itemize}
    If the parameterised network function $\pi \circ f$ is smooth, the projections space is $(0,1)^d \subseteq \bR^d$, and the parameters $\theta$ are found to maximize the mutual information $\mi(Z_1;Z_2)$ lower bound~\eqref{eq:base_decomp} with the reconstruction density $\sfq_{Z_2|Z_1=z_1}(z_2) = C_\textnormal{gauss}(1) e^{- \lVert z_2 - z_2 \rVert_2^2}$, then there is a bijection between the projections $Z$ and the true semantic variables ${S}$.
\end{theorem}

\begin{proof}
    The proof follows directly by~\citet[Theorem~4.4]{von2021self} by noting that the maximizing of the mutual information lower bound~\eqref{eq:base_decomp} with the reconstruction density $\sfq_{Z_2|Z_1=z_1}(z_2) = C_\textnormal{gauss}(1) e^{- \lVert z_2 - z_2 \rVert_2^2}$ coincides with the minimisation of their theorem.
\end{proof}

\section{Algorithm}
\label{app:algorithm }
\Cref{alg:main} describes the main algorithm to maximise the \ba bound. The algorithm includes the possibility of considering the projections in the standard projection space $\cZ$, which usually is the $d$-shpere, or to further generate discrete surrogates in $\{1, 2, \ldots, d\}$. In case the projections are not further processed, it allows to use either \citet{joe1989estimation}'s or the plug-in KDE estimators for the entropy. Finally, it also includes an option to add the \ba bound into distillation methods. The algorithm does not spell out how to extend it to more than two views, but that is done in the usual way, see e.g.~\citep{caron2021emerging}.

\begin{figure*}[p]
    \centering
    \begin{minipage}[t]{.48\textwidth}
        \centering
        \includegraphics[width=0.8\linewidth]{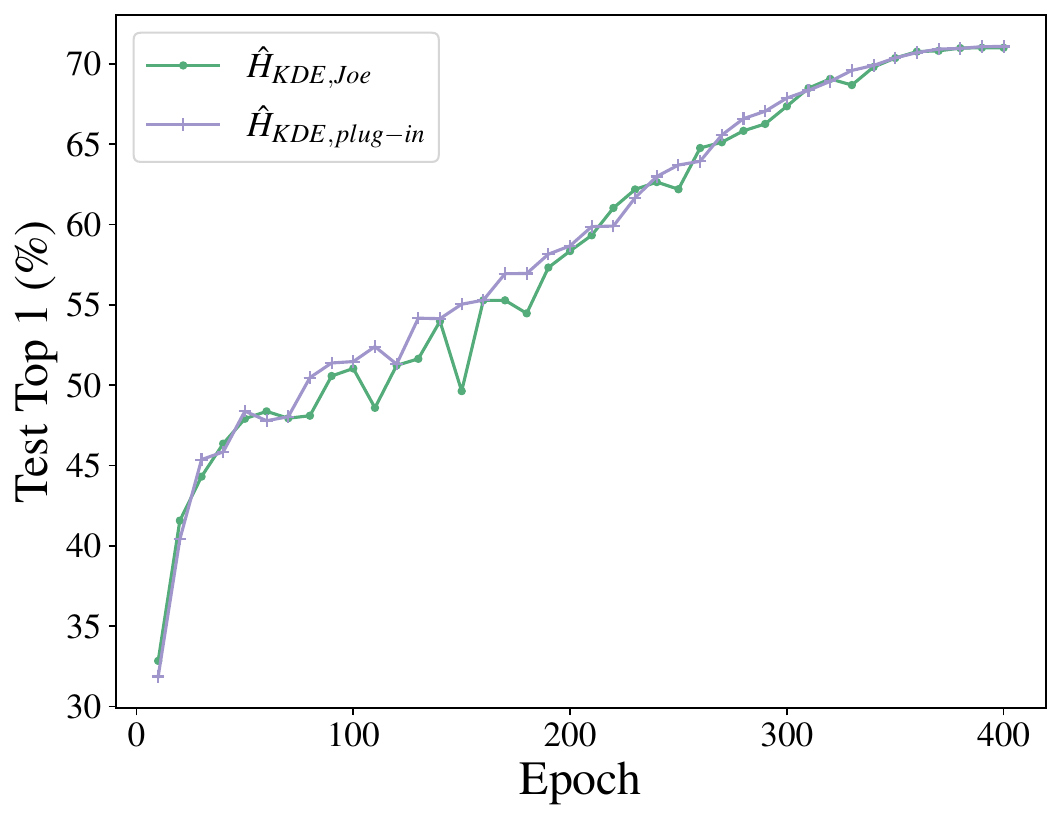}
        \caption{\textit{Joe's and the plug-in KDE achieve the same performance when used to optimize the \er bound.} \simclr + \er top-1 test accuracy on the ImageNet with different KDE to estimate the entropy term.}
        \label{fig:kde_comparison}
    \end{minipage}%
    \hfill
    \begin{minipage}[t]{0.48\textwidth}
        \centering
        \includegraphics[width=0.8\linewidth]{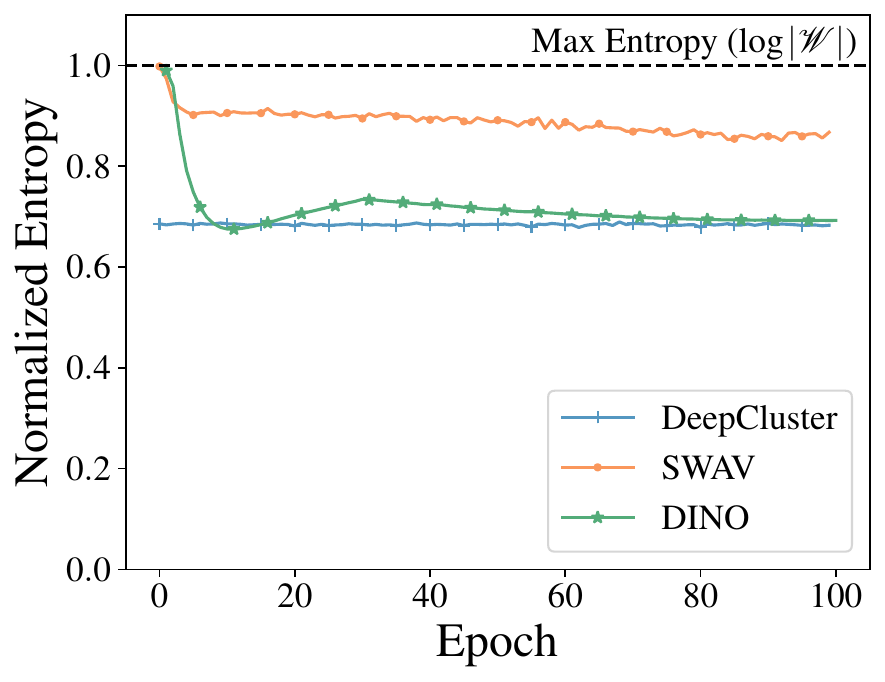}
        \caption{\textit{\dino, \swav, and \deepcluster keep a high entropy in their embedding spaces}.}
        \label{fig:swav_dino_ent}
    \end{minipage}
\end{figure*}

\begin{algorithm*}[p]
\caption{MVSSL algorithm maximising the \ba bound}
\label{alg:main}

\textbf{Input:} Dataset $\cD = \ind{x}{1:n}$, batch size $k$, flag for discrete surrogate \texttt{is\_discrete}, reconstruction density $\sfq_{\textnormal{rec}}$, kernel density $\sfq_{\textnormal{KDE}}$, flag for \citet{joe1989estimation}'s or plug-in estimators \texttt{is\_Joe}, encoder and projector networks $f_\theta$ and $\pi_\theta$, flag for distillation or not distillation \texttt{is\_distillation}, EMA parameter $\lambda \in (0,1)$, learning rate $\eta$, learning rate schedule, augmentation set $\cT$, and number of iterations \texttt{niter}.

\begin{algorithmic}
    \STATE Set \texttt{iter} = 1.
    \WHILE {\texttt{iter} $\leq$ \texttt{niter}}
        \STATE Draw a batch $\ind{x}{1:k}$ uniformly at random from the dataset $\cD$.
        \STATE Draw two sets of augmentations $\ind{t}{1:k}$ and $\ind{t}{1:k}$ uniformly at random from $\cT$.
        \FOR{all $i \in \{1, \ldots, k\}$}
            \STATE Retrieve the projections $\ind{z_1}{i} = \pi_\theta \circ f_\theta \circ \ind{t_1}{i} (\ind{x}{i})$ and $\ind{z_2}{i} = \pi_\theta \circ f_\theta \circ \ind{t_2}{i} (\ind{x}{i})$.
        \ENDFOR
        \IF{\texttt{is\_distillation}}
            \STATE Apply a stop gradient to the teacher projections: $\ind{z_2}{i} \leftarrow \texttt{stopgrad}(\ind{z_2}{i})$.
        \ENDIF
        \IF{\texttt{is\_discrete}}
            \STATE Calculate the empirical p.m.f.s $\hat{\sfp}_1 = \frac{1}{k} \sum_{i=1}^k \sfs(\ind{z_1}{i})$ and $\hat{\sfp_2} = \frac{1}{k} \sum_{i=1}^k \sfs(\ind{z_2}{i})$.
            \STATE Calculate the entropy estimators $\hat{\ent}_1$ and $\hat{\ent}_2$ using the empirical p.m.f.s according to~\eqref{eq:ent_plug_in_disc}
            \STATE Calculate the reconstruction estimates $\widehat{\textnormal{Rec}}_1$  and $\widehat{\textnormal{Rec}}_2$ according to~\eqref{eq:estimate_reconstruction_dist}.
        \ELSE
            \STATE Calculate the densities' KDE $\hat{\sfp}_1$ and $\hat{\sfp}_2$ according to~\eqref{eq:kde_density}.
            \IF{\texttt{is\_Joe}}
                \STATE Calculate~\citet{joe1989estimation}'s KDE of the entropy $\hat{\ent}_1$ and $\hat{\ent}_2$ using the densities' KDE according to~\eqref{eq:entropy_estimate}.
            \ELSE 
                \STATE Calculate the plug-in KDE of the entropy $\hat{\ent}_2$ and $\hat{\ent}_2$ using the densities' KDE according to~\eqref{eq:ent_plug_in_cont}.
            \ENDIF
            \STATE Calculate the reconstruction estimates $\widehat{\textnormal{Rec}}_1$  and $\widehat{\textnormal{Rec}}_2$ according to~\eqref{eq:estimate_reconstruction_cont}.
        \ENDIF
        \STATE Calculate the loss $\cL = - 0.5 \big( \hat{\ent}_1 + \widehat{\textnormal{Rec}}_1 + \hat{\ent}_2 + \widehat{\textnormal{Rec}}_2 \big)$.
        \STATE Update the network weights using gradient descent: $\theta \leftarrow \theta - \eta \nabla_\theta \cL$.
        \IF{\texttt{is\_distillation}}
            \STATE Update teacher's weights according to $\xi \leftarrow \lambda \xi + (1-\lambda) \theta$.
        \ENDIF
        
    \ENDWHILE

\end{algorithmic}

\end{algorithm*}

\begin{figure*}[t]
\centering
\includegraphics[width=0.8\linewidth]{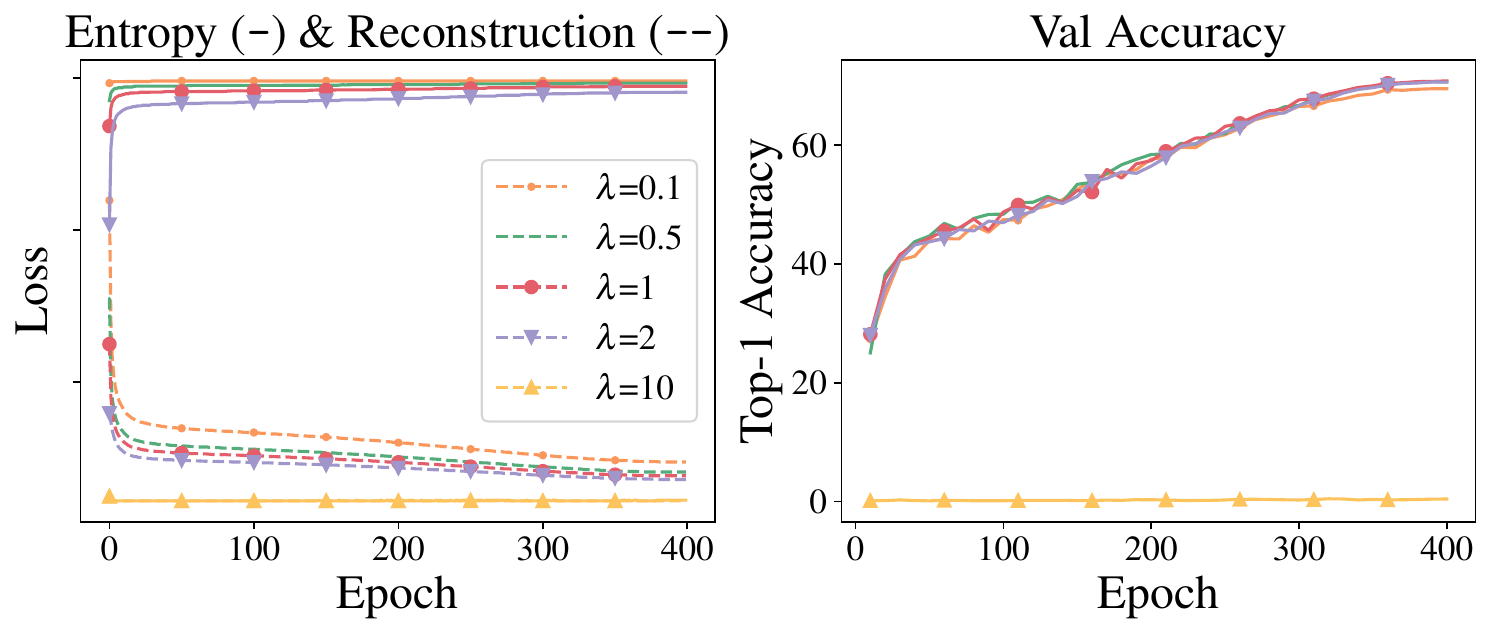}
\caption{\textit{Increasing the weight ($\lambda$) of the reconstruction term leads to poor performance.} Left: evolution of the entropy (\texttt{-}) and reconstruction (\texttt{--}) of \er embedding spaces. Right: top-1 validation accuracy (\%) on ImageNet.}
\label{fig:reconstruction_w_plot}
\end{figure*}

\section{Performance of different KDE estimators}
\label{app:differnt_continuous_estimators}
In \Cref{subsec:mi_between_projections}, we show that the entropy term of \er can be approximated with \citet{joe1989estimation}'s KDE or the plug-in estimator~\citep{krishnamurthy2015lectures}. In \Cref{fig:kde_comparison} we empirically show that using both estimators to optimize the \er bound on \simclr (\simclr + \er) leads to the same performance in terms of ImageNet top-1 accuracy.

\section{Extended results}
\label{app:extended_results}
\Cref{tab:app_imnet_acc} extends \Cref{tab:imnet_acc} with \cmc and EMA ablation for \simclr + \er and \dino + \er. As it can be seen the EMA is essential for distillation methods to achieve good performance. Thus, we hypothesize that the function of EMA is not only limited to keeping the entropy high. We observe that all contrastive methods, including \cmc, are more stable when reducing the batch size than distillation methods. 

\looseness=-1 For \Cref{tab:app_imnet_acc}, we used a continuous entropy KDE-estimator for every method. This includes \dino, contrary to the exact discrete estimator used in the main text (\Cref{tab:imnet_acc}), showing that \er is resilient to the entropy estimator used. To make it clearer, we run an ablation with the two estimators and observe no significant difference in their results (see \Cref{tab:app_imnet_acc_diff_kde_dino}).

\begin{table}[ht]

\caption{\textit{Training w/ \ba yields comparable performance while it improves robustness to changes in batch size.} \textbf{Model}: the set of augmentations, loss, and projection head. $^*$Our re-implementation of the original. \textbf{\er}: the original loss has been substituted by the \ba bound~\eqref{eq:base_decomp}. \textbf{EMA}: the model uses EMA. \textbf{MI}: known to maximize MI. Parentheses: no formal proof provided~(\Cref{subsec:dealing_with_ema}). $\mathbf{\Delta 512}$: accuracy drop when reducing batch size to 512.}
\label{tab:app_imnet_acc}
\begin{tabular}{lcccc}
\toprule
 Model &                 EMA &    MI    &Acc ($\uparrow$)     &  $\Delta 512$ ($\downarrow$) \\
\midrule
  \dino{}$^*$ &     $\checkmark$ & ?                 &   75.28 &   \textcolor{white}{0}8.63 \\
  \dino + \ba &     $\times$ &    $\checkmark$      &    67.30 &    \textcolor{white}{0}5.45 \\
  \dino{} + \ba &  $\checkmark$ &    $(\checkmark)$      &  73.63 &   \textcolor{white}{0}2.67 \\\hline
  \byol{}$^*$ & $\checkmark$ & ?                 &   73.42 &   23.65 \\ 
  \byol + \ba &     $\times$ &    $\checkmark$      &    71.70 &    \textcolor{white}{0}3.20 \\
  \byol{} + \ba & $\checkmark$ &    $(\checkmark)$      &  71.94 &   \textcolor{white}{0}2.35 \\\hline
  \cmc{}$^*$ &      $\times$ &    $\checkmark$      &    69.95 &   \textcolor{white}{0}3.06 \\
  \simclr{}$^*$ &  $\times$ &    $\times    $      &    70.23 &   \textcolor{white}{0}2.17 \\
  \simclr{} + \ba &     $\times$ &$ \checkmark $    &    70.86 &   \textcolor{white}{0}1.01 \\
\bottomrule
\end{tabular}
\end{table}

\begin{table}[ht]
\centering
\caption{\textit{There is no significant difference between the discrete and the continuous entropy estimators.}}
\label{tab:app_imnet_acc_diff_kde_dino}
\begin{tabular}{lccc}
\toprule
Model & Discrete & Acc ($\uparrow$) & $\Delta 512$ ($\downarrow$)\\ \midrule
  \dino{}$^*$ &     $\checkmark$ &   75.28 &   8.63 \\\midrule
    \dino + \er &  $\checkmark$ & 73.39 &  2.35 \\
  \dino{} + \ba &  $\times$ &   73.63 &   2.67\\ \bottomrule
\end{tabular}
\end{table}

\section{Entropy minimization in discrete MVSSL algorithms}
\label{app:discrete_mvssl}
\Cref{fig:swav_dino_ent} shows the evolution of the entropy for discrete methods such as  \swav, \deepcluster, and \dino. The entropy is normalized by $\log |\mathcal{W}|$, which is the maximum possible entropy for these methods. We observe that while the normalized entropy does not increase over time, its value is kept close to the maximum. This is expected for \swav and \deepcluster, since we have shown that they maximize the \er~(\Cref{susec:clustering_methods}). For \dino, we showed that softmax centering and sharpening could be used to maintain a high entropy value (\Cref{subsec:distillation_methods} and \Cref{fig:swav_dino_ent}).

The behavior of the entropies was to be expected, as we discuss below.

In \swav, the Sinkhorn-Knopp in the first iteration almost completely accomplishes that the conditional entropy $\sfp_{W_2|Z_2} = \bm{1}_m / m$, as it has full liberty to do so. Therefore, the marginal entropy $\sfp_{W_2}$ is uniform and the entropy is maximal. As the iterations continue, the tension between the cross entropy related to the reconstruction term and the entropy maximization, together with the fact that \citet{caron2020unsupervised} only perform three iterations of the Sinkhorn-Knopp algorithm, push the conditional entropy $\sfp_{W_2|Z_2}$ slightly away from uniformity, thus decreasing the entropy.

In \deepcluster, the images of each batch are sampled based on a uniform distribution over cluster assignments, i.e. for each batch $\sfp_{W_2} \approx \frac{1}{k} \sum_{i=1}^k \sfp_2^{(i)}$ is almost uniform. This way, the entropy $\ent(W_2)$ is approximately maximized.

In \dino, the centering step at the first iteration, before the weights have been updated, completely accomplishes a uniform conditional distribution, as then $\xi = \theta$ and thus the sharpening has no effect. In the second iteration, the weights already are different $\xi \neq \theta$, and the low value of the temperature pushes the conditional entropy $\sfp_{W_2|Z_2}$ away from uniformity. This is compensated by the centering, which avoids that $\sfp_{W_2|Z_2}$ becomes completely degenerate. Therefore, the result is an entropy that is lower than the maximal (sharpening) but not minimal (centering). The tension between these to mechanisms evolves towards higher entropies as the temperature is scheduled to increase over time, and thus the entropy does too.

\section{Additional experiments with different weights for the reconstruction objective}
\label{app:different_weights}

Recall the \er objective from~\eqref{eq:base_decomp}. This objective is theoretically justified as a lower bound on the mutual information between the projections of different branches. Moreover, \Cref{sec:background_and_related_work} gives an intuition of the role of both the entropy and the reconstruction terms. Empirically, it may be interesting to consider giving different weights to each of the terms and see if this may lead to better performance. More precisely, it  may be interesting to consider the objective
\begin{equation*}
    \mi_{\er,\lambda} \coloneqq \ent(Z_2) + \lambda \ \bE[\log \sfq_{Z_2|Z_1}\!(Z_2)]
\end{equation*}
for different values of $\lambda > 0$ as we show in \Cref{fig:reconstruction_w_plot}.

Intuitively, for projections in $\bR^d$ that are not further projected into a discrete surrogate, the reconstruction is given by~\eqref{eq:estimate_reconstruction_cont} choosing the reconstruction density $q_{Z_2|Z_1=z_1}(z_2) \propto \exp f(z_2,z_1)$, where $f(z_2,z_1)$ is a function measuring the similarity of $z_1$ and $z_2$. Therefore, the objective $\mi_{\er,\lambda}$ is equivalent to consider a density $q_{Z_2|Z_1=z_1}(z_2) \propto \exp \lambda f(z_2,z_1)$. Under this interpretation, we may understand better the results obtained with the different weights since lower values of $\lambda$ lead to flatter densities and high values of $\lambda$ to very spikier densities: 
\begin{itemize}
    \item If the density is very flat, then the reconstruction term does not vary much even if $Z_1$ and $Z_2$ are close. This happens as the reconstruction density makes $Z_1$ and $Z_2$ only loosely dependent.  Therefore, the network has a hard time learning and the performance decreases.
    \item If the density is very spiky (almost a delta), then it means that $Z_1 \approx Z_2$, so the network collapses. 
\end{itemize}

When the projections are projected into a discrete surrogate, then the $\lambda$ parameter should only be interpreted as a temperature parameter similar to those in other MVSSL methods. However, the understanding and intuition provided above still apply.

\end{document}